\documentclass[journal]{IEEEtran}

\let\labelindent\relax
\usepackage[utf8]{inputenc}

\usepackage{cite}

\usepackage{todonotes}
\usepackage{changes}
\usepackage[most]{tcolorbox} 
\usepackage{subcaption}
\usepackage{booktabs}
\usepackage{svg}
\usepackage{csquotes}
\usepackage{multicol}
\usepackage{amsmath, amssymb}
\usepackage{multirow}
\usepackage{float}
\usepackage{enumitem}
\usepackage{makecell}
\usepackage[english]{babel}
\usepackage{hyphenat}
\usepackage{hyperref}
\usepackage{textcomp}

\newcommand{\aref}[1]{\hyperref[#1]{Appendix~\ref*{#1}}}

\usepackage{balance}

\begin{document}
\title{SkyDreamer: Interpretable End-to-End Vision-Based Drone Racing with Model-Based Reinforcement Learning}
\author{Aderik Verraest, Stavrow Bahnam, Robin Ferede, Guido de Croon, Christophe De Wagter
\thanks{The authors are with the Micro Air Vehicle Lab of the Faculty of Aerospace Engineering, Delft University of Technology, 2629 HS Delft, The Netherlands}}


\maketitle

\begin{abstract}
Autonomous drone racing (ADR) systems have recently achieved champion-level performance, yet remain highly specific to drone racing. While end-to-end vision-based methods promise broader applicability, no system to date simultaneously achieves full sim-to-real transfer, onboard execution, and champion-level performance. In this work, we present \textit{SkyDreamer}, to the best of our knowledge, the first end-to-end vision-based ADR policy that maps directly from pixel-level representations to motor commands. SkyDreamer builds on \textit{informed Dreamer}, a model-based reinforcement learning approach where the world model decodes to privileged information only available during training. By extending this concept to end-to-end vision-based ADR, the world model effectively functions as an implicit state and parameter estimator, greatly improving interpretability. SkyDreamer runs fully onboard without external aid, resolves visual ambiguities by tracking progress using the state decoded from the world model’s hidden state, and requires no extrinsic camera calibration, enabling rapid deployment across different drones without retraining. Real-world experiments show that SkyDreamer achieves robust, high-speed flight, executing tight maneuvers such as an inverted loop, a split-S and a ladder, reaching speeds of up to $21 \, \text{m/s}$ and accelerations of up to $6 \, \text{g}$. It further demonstrates a non-trivial visual sim-to-real transfer by operating on poor-quality segmentation masks, and exhibits robustness to battery depletion by accurately estimating the maximum attainable motor RPM and adjusting its flight path in real-time. These results highlight SkyDreamer’s adaptability to important aspects of the reality gap, bringing robustness while still achieving extremely high-speed, agile flight.

\end{abstract}


\setlist[itemize]{noitemsep}

\section{Introduction}

Recent advancements in autonomous drone racing (ADR) have reached champion-level performance, even beating human world champions \cite{Kaufmann2023, aibeatshumandeWagter2025}. In 2023, Kaufmann \textit{et al.} \cite{Kaufmann2023} became the first to defeat human champion drone racers with autonomous drones. More recently, Bahnam \textit{et al.} \cite{aibeatshumandeWagter2025} achieved the first champion-level victory in an externally organized competition, relying solely on a low-cost monocular rolling shutter camera for perception.

\begin{figure}[t!]
    \centering
    \begin{subfigure}[b]{\columnwidth}
        \centering
        \includegraphics[width=1.01\columnwidth]{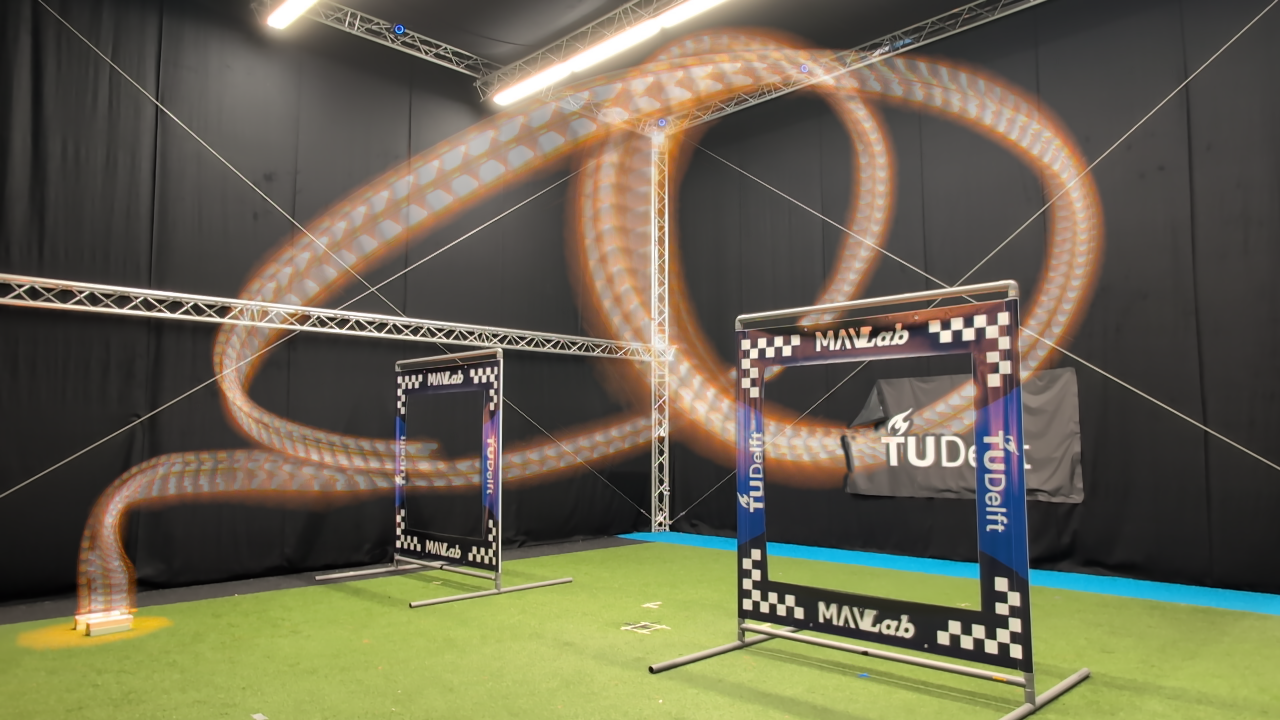}
    \end{subfigure}
    
    \vspace{1.0em}
    
    \begin{subfigure}[t]{0.65\columnwidth}
        \raggedright
        \includegraphics[height=3cm]{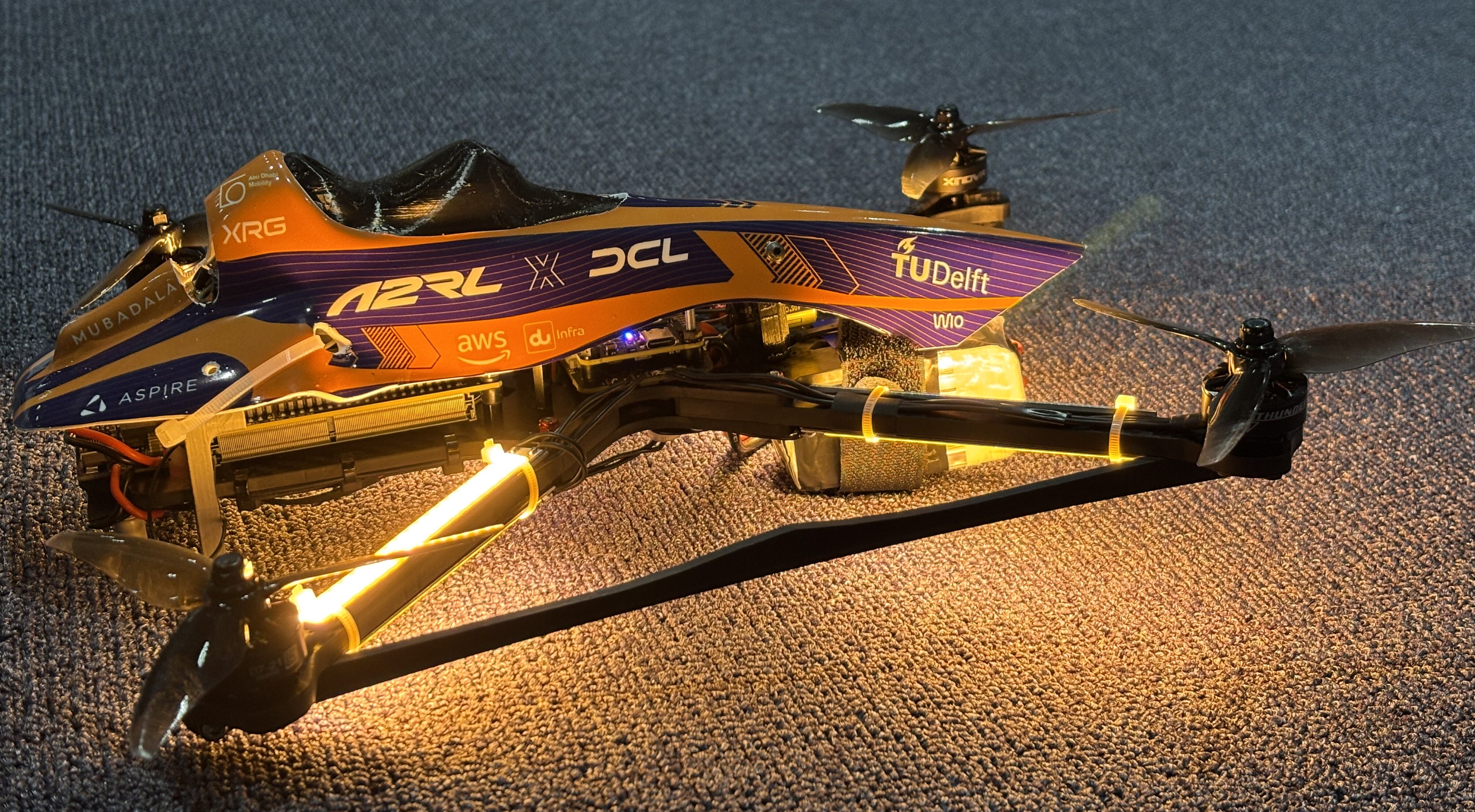}
    \end{subfigure}
    \hfill
    \begin{subfigure}[t]{0.33\columnwidth}
        \raggedleft
        \includegraphics[height=3cm]{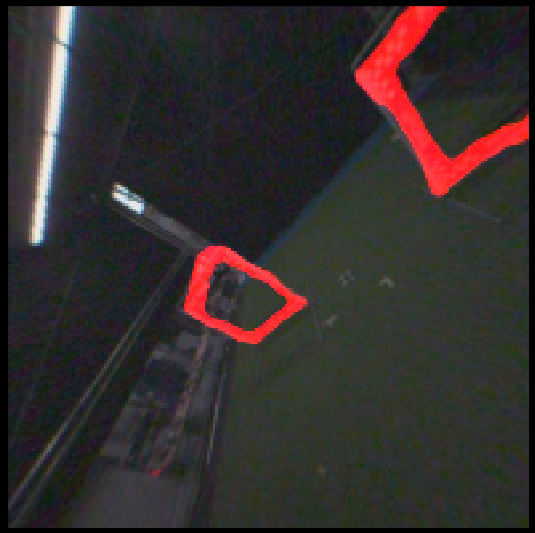}
    \end{subfigure}
    \caption{Real-world inverted loop track with MAVLab gates. SkyDreamer flies agile maneuvers such as an inverted loop in the real world without external aid (top), operates end-to-end on poor-quality segmentation masks obtained from onboard images (shown in red, bottom-right), and directly controls the physical platform by outputting low-level motor commands. In a flight area confined to $6 \times 6 \, \text{m}$, it reaches speeds of up to $13 \, \text{m/s}$ and accelerations of up to $6 \, \text{g}$. The racing drone is identical to the one used in the A2RL x DCL Autonomous Drone Championship 2025 \cite{aibeatshumandeWagter2025} (bottom-left).}
    \label{fig:overview_image}
\end{figure}

Despite these successes, current champion-level ADR systems \cite{aibeatshumandeWagter2025, Kaufmann2023} remain modular pipelines of classical algorithms. They are highly specialized and inflexible, relying on gate corner detection and Perspective-n-Point (PnP) for state estimation. Their performance further depends on accurately mapped track layouts, rectangular gates of known size, precise camera calibration, and hand-tuned extended Kalman filters (EKFs). Optimizing perception, state estimation, and control in isolation further limits overall performance. Such pipelines cannot generalize beyond the highly structured settings they were designed for. In contrast, humans fly with remarkable adaptability -- handling unseen tracks, different drones, and even unstructured environments with minimal task-specific training \cite{scaramuzza_human}. This gap underscores the inherent limitations in today’s champion-level ADR systems.

\begin{table*}[ht]
\centering
\caption{Comparison of recent end-to-end vision-based ADR approaches with SkyDreamer. “Offboard” indicates that some computation occurs offboard, while “onboard” means all computation runs onboard with no external aid during flight. “HIL” denotes that images for the policy during real-world flight are rendered using MoCap. “Visual ambiguity” refers to situations where visually similar observations require different flight paths, such as in the ladder inverted loop track shown in \autoref{fig:perception_ladder_sim_combined}. The perception reward encourages the drone to look toward gates, introducing human bias. Only performance metrics from real-world experiments are considered in this comparison.}
\begin{tabular}{lccccc}
\toprule
\textbf{Criterion} & Geles \textit{et al.} \cite{geles2024demonstratingagileflightpixels} & Xing \textit{et al.} \cite{xing2024bootstrappingreinforcementlearningimitation} &  Romero \textit{et al.} \cite{romero2025dreamflymodelbasedreinforcement} & Krinner \textit{et al.} \cite{krinner2025acceleratingmodelbasedreinforcementlearning} & \textbf{SkyDreamer (ours)} \\
\midrule
\makecell[l]{\textbf{Onboard} \\ \textbf{computation}} & Offboard & Offboard & Offboard & Offboard & \makecell{\textbf{Onboard} \\ \textbf{without external aid}} \\
\textbf{Champion-level} & No ($2\,g$) & No ($2.7\,\text{TWR}$) \footnotemark & No ($2.7\,\text{TWR}$) & No & Likely ($6 \; g$) \\
\makecell[l]{\textbf{Sim-to-real} \\ \textbf{transfer}} & \textbf{Yes} & HIL (only dynamics) & HIL (only dynamics) & HIL (only dynamics) & \makecell{\textbf{Yes, non-trivial} \\ \textbf{visual sim-to-real gap}} \\
\textbf{Interpretability} & None & None & \makecell{Limited, latent $\rightarrow$ \\ reconstructed image} & \makecell{\textbf{Yes, via decoding to} \\ \textbf{states}} & \makecell{\textbf{Yes, via decoding to} \\ \textbf{states and parameters}} \\
\makecell[l]{\textbf{Robust to} \\ \textbf{visual ambiguity}} & Not demonstrated & Not demonstrated & Not demonstrated & Not demonstrated & \makecell{\textbf{Yes, via progress tracking} \\ \textbf{using world model}} \\
\makecell[l]{\textbf{Perception} \\ \textbf{reward}} & Yes & Yes & \textbf{No, emergent behavior} & \textbf{No, emergent behavior} & \textbf{No, emergent behavior} \\
\bottomrule
\end{tabular}
\label{tab:adr_comparison}
\end{table*}

To address these limitations, research has turned toward end-to-end vision-based learning, where policies map pixel-level representations directly to actions without explicit state estimation \cite{romero2025dreamflymodelbasedreinforcement, geles2024demonstratingagileflightpixels, xing2024bootstrappingreinforcementlearningimitation, krinner2025acceleratingmodelbasedreinforcementlearning}. Such learned representations offer greater flexibility, potentially enhancing performance by jointly optimizing perception, state estimation, and control. Crucially, leveraging pixel-level inputs, such as depth maps, can generalize to unstructured environments while introducing a smaller reality gap compared to raw RGB images, as demonstrated in \cite{high_speed_wild, mavrl, yu2025depthtransferlearninglike, learning_vision_based_agile_flight_diff_physics}. At the same time, end-to-end learning reduces reliance on human-designed abstractions and reward shaping, which can constrain system performance.

\footnotetext{Xing \textit{et al.} \cite{xing2024bootstrappingreinforcementlearningimitation} did perform limited experiments in simulation demonstrating champion-level performance comparable to \cite{Kaufmann2023}, but failed to demonstrate such performance in the real world.}

While these benefits are promising, practical implementations remain challenging. Only Geles \textit{et al.} \cite{geles2024demonstratingagileflightpixels} manage to bridge the visual reality gap, whereas Xing \textit{et al.} \cite{xing2024bootstrappingreinforcementlearningimitation}, Romero \textit{et al.} \cite{romero2025dreamflymodelbasedreinforcement} and Krinner \textit{et al.} \cite{krinner2025acceleratingmodelbasedreinforcementlearning} still rely on hardware-in-the-loop (HIL) simulation, where images during real-world flights are rendered using an external motion capture (MoCap) system rather than the real camera feed. Furthermore, as summarized in \autoref{tab:adr_comparison}, none of these approaches perform all computation fully onboard, nor do they demonstrate champion-level performance in the real world, as the thrust-to-weight ratio (TWR) was deliberately software-capped. To date, no end-to-end vision-based method simultaneously achieves full sim-to-real transfer, onboard execution, and champion-level flight.

An additional, and often overlooked, constraint is that most ADR approaches treat collective thrust and body rates (CTBR) as the final control layer, relying on an inner-loop controller such as PID or INDI at deployment to track these CTBR actions with motor commands \cite{Kaufmann2023, geles2024demonstratingagileflightpixels, romero2025dreamflymodelbasedreinforcement, xing2024bootstrappingreinforcementlearningimitation, krinner2025acceleratingmodelbasedreinforcementlearning}. As an alternative to this two-stage control structure, the European Space Agency (ESA) introduced guidance and control networks (G\&CNETs), which map state inputs directly to actuator commands \cite{origer2023guidancecontrolnetworks}. For quadcopter control, this translates to generating motor commands directly rather than relying on intermediate CTBR actions. Doing so is required for several space applications and simplifies modeling, as the motor response is generally easier to capture than the complex behavior of an inner-loop controller. Additionally, direct motor control may also enable more optimal performance compared to approaches that rely on inner-loop controllers.

Building on this concept, Ferede \textit{et al.} \cite{robinrlfirstrlone} trained G\&CNETs using reinforcement learning (RL), specifically proximal policy optimization (PPO) \cite{ppo}, for ADR, demonstrating that fully neural control policies can outperform traditional cascaded controllers when trained under comparable conditions \cite{supervised_nn_control_robin, robinrlfirstrlone, ferede2025netrulealldomain}. The use of G\&CNETs for quadcopter control has already proven successful in ADR competitions: during the A2RL x DCL Autonomous Drone Championship in 2025, a neural end-to-end state-based controller outperformed human world-champion pilots, underscoring its real-world potential \cite{aibeatshumandeWagter2025}.

A promising approach to training end-to-end vision-based ADR policies is model-based RL. Unlike model-free RL methods, such as PPO \cite{ppo} and soft actor-critic (SAC) \cite{sac}, model-based RL learns an internal model of the environment, often referred to as a \textit{world model} \cite{world_models_original}. The core idea is to capture environment dynamics by learning to predict future observations -- or their latent representations -- based on past observations and actions using the world model. Actor training is thus performed entirely in latent space, where the actor interacts with the learned world model rather than the physical or simulated environment. This is accomplished through latent imagination, in which future latent states are predicted from previous latent states and actions. Consequently, sample efficiency is greatly improved: physical or simulated rollouts are required only to train the world model, not the actor. This results in orders-of-magnitude gains in sample efficiency compared to model-free RL, which is particularly important for end-to-end vision-based policies, where rendering observations at every timestep can be computationally expensive \cite{world_models_original, dreamerv3_nature}.

In model-based RL, the Dreamer algorithms introduced by Hafner \textit{et al.} have become among the most widely adopted \cite{dreamerv3_nature, planet, dreamerv1, dreamerv2, wu2022daydreamerworldmodelsphysical}. Notably, Wu \textit{et al.} \cite{wu2022daydreamerworldmodelsphysical} demonstrated that a robot could be trained to walk using only one hour of real-world interactions -- without relying on simulation -- highlighting the remarkable sample efficiency of model-based RL. More recently, DreamerV3 has emerged as a particularly promising variant, achieving state-of-the-art performance across more than 150 diverse environments using a single set of hyperparameters \cite{dreamerv3_nature}. Romero \textit{et al.} \cite{romero2025dreamflymodelbasedreinforcement} have already applied DreamerV3 to end-to-end vision-based ADR, though with limited success both in terms of performance and sim-to-real transfer. 

To further aid learning efficiency, the use of privileged information during training has shown great promise \cite{geles2024demonstratingagileflightpixels, xing2024bootstrappingreinforcementlearningimitation}. In model-based RL, this idea can be extended by allowing the world model to decode privileged observations, such as ground-truth states, an idea already explored for end-to-end vision-based ADR in Krinner \textit{et al.} \cite{krinner2025acceleratingmodelbasedreinforcementlearning}. This has been shown to yield significant performance gains, particularly in tasks where such states are recoverable from sequences of observations \cite{lambrechts2024informedpomdpleveragingadditional, hu2024privilegedsensingscaffoldsreinforcement}, as in vision-based ADR. Furthermore, privileged reconstructions enable the world model to act as an implicit state estimator, improving interpretability, unlike prior approaches \cite{geles2024demonstratingagileflightpixels, romero2025dreamflymodelbasedreinforcement, xing2024bootstrappingreinforcementlearningimitation} that largely operate as black boxes.

Alternative model-based approaches such as TD-MPC2 combine latent imagination with model predictive control (MPC) \cite{hansen2024tdmpc2scalablerobustworld}. By relying on MPC rather than RL, TD-MPC2 outperforms DreamerV3 on several continuous control benchmarks \cite{hansen2024tdmpc2scalablerobustworld}. However, unlike Dreamer, TD-MPC2 lacks a recurrent structure in latent space and is therefore unable to model long-range temporal dependencies. Its reliance on iterative optimization at runtime further limits its suitability for onboard deployment. Other latent-space MPC approaches include \cite{s2023gradientbasedplanningworldmodels} and \cite{zhou2025dinowmworldmodelspretrained}. Notably, Zhou \textit{et al.} \cite{zhou2025dinowmworldmodelspretrained} demonstrated zero-shot planning with a pretrained billion-parameter world model. While impressive, the scale of such models makes deployment on resource-constrained platforms like drones impractical.

\noindent\emph{Contributions}

In this work, we present \textbf{SkyDreamer}, to the best of our knowledge the first end-to-end vision-based ADR policy that maps \textbf{directly from pixels to motor commands}, eliminating the need for inner-loop controllers in end-to-end vision-based drone racing. Beyond this main contribution, our work offers several further advances, summarized below and in \autoref{tab:adr_comparison}:
\begin{itemize}[leftmargin=*]
    \item SkyDreamer greatly improves \textbf{interpretability} by decoding privileged information, effectively turning the world model into an implicit state and parameter estimator,
    \item is, to the best of our knowledge, the first end-to-end vision-based ADR policy to reach \textbf{accelerations of up to $\pmb{6} \, \text{g}$ and speeds of up to $\pmb{21} \, \text{m/s}$} in the real world, on par with state-of-the-art classical approaches \cite{aibeatshumandeWagter2025, Kaufmann2023},
    \item demonstrates non-trivial \textbf{visual sim-to-real transfer} using poor-quality segmentation masks,
    \item \textbf{resolves visual ambiguity} by explicitly tracking progress through the state decoded from the world model’s hidden state, laying the foundation for end-to-end vision-based policies capable of flying arbitrary tracks,
    \item requires \textbf{no accelerometer},
    \item runs \textbf{fully onboard without external aid},
    \item \textbf{estimates drone-specific parameters, such as camera extrinsics, on the fly} and operates on images standardized via intrinsic calibration, enabling rapid deployment across different drones without retraining.
\end{itemize}

\section{Methodology}

\subsection{Problem statement}\label{subsec:problem_statement}
We aim to develop an end-to-end vision-based policy capable of flying a quadcopter at high speed through a known track using only onboard computation and sensing. Specifically, the goal is to maximize the quadcopter's lap speed around a pre-mapped track marked by gates, subject to the constraints of the physical platform and the requirement to pass through all gates in a predefined order while avoiding collisions. The policy receives only raw sensory inputs: pixel-level representations, body rate measurements, and motor RPM measurements. Additionally, the drone’s camera extrinsic parameters are assumed to be uncalibrated, as extrinsic calibration is both time-consuming and must often be repeated during operation due to changes in the camera angle from touching, crashing, or hitting gates, making it impractical for real-world competitions. The policy must therefore learn to handle this uncertainty. It shall map these observations directly to motor commands, without relying on an intermediate inner-loop controller such as PID or INDI. During training only, the policy is granted access to privileged information, including ground-truth states, camera parameters, and dynamic parameters of the system.

We formalize this problem as an \emph{informed partially observable Markov decision process} (informed POMDP) \cite{lambrechts2024informedpomdpleveragingadditional}, which extends the standard POMDP \cite{pomdp}. The informed POMDP was first introduced by Lambrechts \textit{et al.} \cite{lambrechts2024informedpomdpleveragingadditional}, who provide a detailed formulation. The key distinction with a standard POMDP is that additional, privileged information is available during training but not at execution time. This additional information can be exploited to improve the training process.

Formally, the informed POMDP is defined as:
\[
\widetilde{\mathcal{P}} = (\mathcal{S}, \mathcal{U}, \mathcal{I}, \mathcal{O}, T, R, \widetilde{\mathcal{I}}, \widetilde{\mathcal{O}}, P, \gamma)
\]
with the following components:

\begin{itemize}[leftmargin=*]
    \item \textbf{state space} $\mathcal{S}$ with states $\mathbf{s}_t$
    \item \textbf{action space} $\mathcal{U}$ with actions $\mathbf{u}_t$
    \item \textbf{information space} $\mathcal{I}$ with information $\mathbf{i}_t$ only available during training
    \item \textbf{observation space} $\mathcal{O}$ with observations $\mathbf{o}_t$
    \item \textbf{transition distribution} $T(\mathbf{s}_{t+1} \mid \mathbf{s}_t, \mathbf{u}_t)$
    \item \textbf{reward function} $r_t = R(\mathbf{s}_t, \mathbf{u}_t)$, which is designed to reflect the task objective and is detailed further in \autoref{subsec:reward_function}.
    \item \textbf{information distribution:} $\widetilde{\mathcal{I}}(\mathbf{i}_t \mid \mathbf{s}_t)$
    \item \textbf{observation distribution:} $\widetilde{\mathcal{O}}(\mathbf{o}_t \mid \mathbf{i}_t)$ available during training and execution
    \item \textbf{initial state distribution:} $P(\mathbf{s}_0)$, which is detailed further in \autoref{tab:randomization_ranges}
    \item \textbf{discount factor:} $\gamma \in [0, 1)$
\end{itemize}

In an Informed POMDP, the states $\mathbf{s}_t$ are never observable. At training time, the informed POMDP does provide the policy access to a history $\mathbf{h}^+_t$:
\begin{align*}
    \mathbf{h}^+_t = (\mathbf{i}_0, \mathbf{o}_0, \mathbf{u}_0, \mathbf{r}_0, ..., \mathbf{i}_{t-1}, \mathbf{o}_{t-1}, \mathbf{u}_{t-1}, \mathbf{r}_{t-1}, \mathbf{i}_t, \mathbf{o}_t)
\end{align*}
consisting of privileged information, observations, actions and rewards. The observations $\mathbf{o}_t$ are assumed to be conditionally dependent on the states $\mathbf{s}_t$, following the Bayesian network $\mathbf{s}_t \rightarrow \mathbf{i}_t \rightarrow \mathbf{o}_t$. To satisfy this assumption, and following Lambrechts \textit{et al.} \cite{lambrechts2024informedpomdpleveragingadditional}, we define the information variable as $\mathbf{i}_t=[\mathbf{o}_t,\mathbf{o}_t^+]^T$, a concatenation of the observations $\mathbf{o}_t$ and privileged observations $\mathbf{o}_t^+$  only available during training (e.g. position $\mathbf{p}$ or velocity $\mathbf{v}$). At execution time, we define the underlying execution POMDP $\widetilde{\mathcal{P}}$ as:
\begin{align*}
    \widetilde{\mathcal{P}} = (\mathcal{S}, \mathcal{U}, \mathcal{O}, T, R, \widetilde{\mathcal{O}}, P, \gamma)
\end{align*}
which provides access to a history $\mathbf{h}_t$:
\begin{align*}
    \mathbf{h}_t = (\mathbf{o}_0, \mathbf{u}_0, ..., \mathbf{o}_{t-1}, \mathbf{u}_{t-1}, \mathbf{o}_t)
\end{align*}
consisting of only observations and actions \cite{lambrechts2024informedpomdpleveragingadditional, romero2025dreamflymodelbasedreinforcement}.

The objective in an informed POMDP is to maximize the expected return in the underlying execution POMDP by learning a policy $\pi_{\pmb{\theta}}(\mathbf{u_t} | \mathbf{h}_t)$, in our case parameterized by neural network parameters $\pmb{\theta}$. The resulting optimal policy $\pi_\theta^*$ is defined by \cite{lambrechts2024informedpomdpleveragingadditional}:
\begin{align*}
    \pi_\theta^* = \underset{\pi_\theta}{\text{argmax}} \;
    \underset{\substack{
        \mathbf{s}_0 \sim P(\cdot) \\ 
        \mathbf{u}_t \sim \pi_\theta(\cdot \mid \mathbf{h}_t) \\ 
        \mathbf{s}_{t+1} \sim T(\cdot \mid \mathbf{s}_t, \mathbf{a}_t) \\ 
        \mathbf{o}_t \sim \widetilde{\mathcal{O}}(\cdot \mid \mathbf{s}_t)
    }}{\mathbb{E}}
    \left[ \sum_{t=0}^\infty \gamma^t R(\mathbf{s}_t, \mathbf{a}_t) \right]
\end{align*}
The optimal policy thus aims to maximize the expected return, i.e., the cumulative sum of discounted rewards, in the execution POMDP through interaction with the informed POMDP at training time. Although privileged information is used during learning, the resulting optimal policy remains independent of it at execution time.

\subsection{Spaces}\label{subsec:spaces}

\begin{figure}[!ht]
    \centering
    \includegraphics[width=0.48\textwidth]{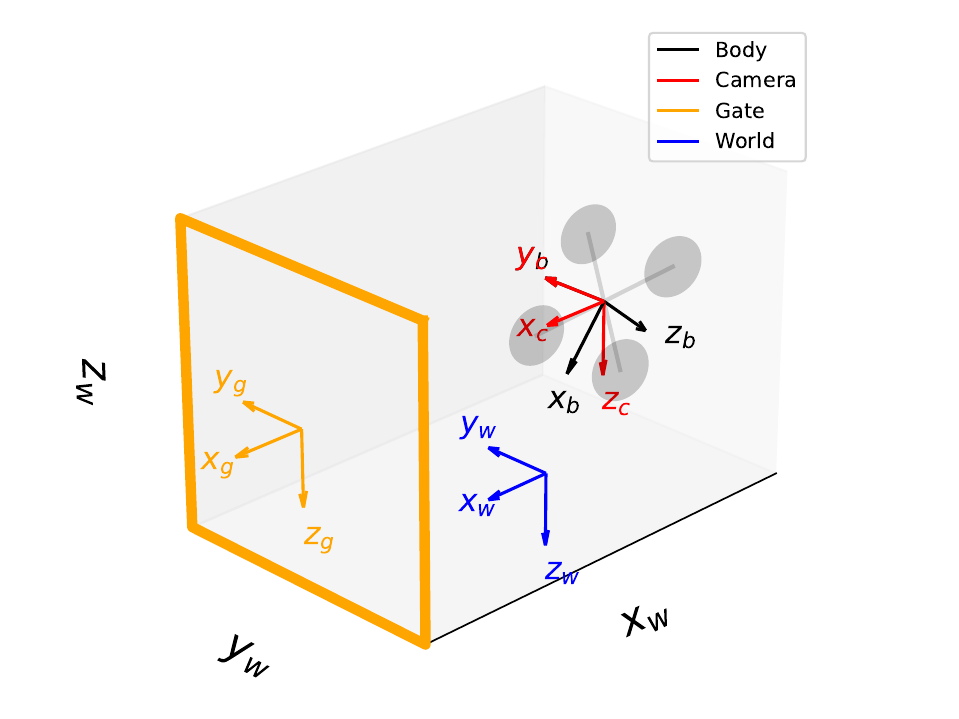}
    \caption{Coordinate systems. A quadcopter pitched forward -- shown as a cross with four circles -- with a camera pitched upward, is shown flying toward the gate at the bottom left. Subscripts indicate the reference frames: $w$ - world frame, $g$ - gate frame, $b$ - body frame, and $c$ - camera frame. Image not to scale, the axes of the frames are not necessarily aligned.}
    \label{fig:coordinate_systems}
\end{figure}

The state space $\mathcal{S}$ consists of states $\mathbf{s}$ which are defined as:
\begin{align*}
\mathbf{s} = [\mathbf{p}_w, \mathbf{p}_g, \mathbf{v}_w, \mathbf{v}_g, \pmb{\lambda}, \pmb{\Omega}, \pmb{\omega}, \mathbf{c}_e, \mathbf{d}, \pmb{\varepsilon}, \mathbf{f}]^\top
\end{align*}
where the associated coordinate systems all follow a north–east–down (NED) convention illustrated in \autoref{fig:coordinate_systems}, and the components of $\mathbf{s}$ are described below:
\begin{itemize}[leftmargin=*]
    \item \textbf{position}:
    \begin{itemize}
        \item $\mathbf{p}_w = [x_w, y_w, z_w]^\top$: position in world frame
        \item $\mathbf{p}_g = [x_g, y_g, z_g]^\top$: position in gate frame, updated upon passing each gate. The gate frame is a local coordinate system attached to the next gate, which is updated each time a gate is passed and is also shown in \autoref{fig:coordinate_systems}
    \end{itemize}
    \item \textbf{velocity:}
    \begin{itemize}
        \item $\mathbf{v}_w = [v_{x,w}, v_{y,w}, v_{z,w}]^\top$: velocity in world frame
        \item $\mathbf{v}_g = [v_{x,g}, v_{y,g}, v_{z,g}]^\top$: velocity in gate frame
    \end{itemize}    
    \item \textbf{orientation} $\pmb{\lambda} = [\phi_w, \theta_w, \psi_w, \psi_g]^\top$ in $\psi \rightarrow \theta \rightarrow \phi$ order
    \begin{itemize}
        \item $\phi_w$, $\theta_w$: roll and pitch in world frame
        \item $\psi_w$: yaw in world frame
        \item $\psi_g$: yaw in gate frame
    \end{itemize}
    \item \textbf{body rates} $\pmb{\Omega} = [p, q, r]^T$ are the roll, pitch and yaw rate in body frame respectively
    \item \textbf{propeller speeds} $\pmb{\omega} = [\omega_1, \omega_2, \omega_3, \omega_4]^T$ are the angular velocities of the four propellers
    \item \textbf{camera extrinsics} $\mathbf{c}_e = [\phi_c, \theta_c, \psi_c]$
    describes the euler rotation from body to camera frame in $\psi \rightarrow \theta \rightarrow \phi$ order
    \item \textbf{dynamic parameters} $\mathbf{d}$ are randomized dynamical parameters of the drone described in \autoref{subsec:dynamics}
    \item \textbf{disturbances} $\pmb{\varepsilon} = [\pmb{\varepsilon}_a, \pmb{\varepsilon}_M, \pmb{\varepsilon}_u]^\top$
    \begin{itemize}
        \item $\pmb{\varepsilon}_a = [\varepsilon_{a,x}, \varepsilon_{a,y}, \varepsilon_{a,z}]^\top$: acceleration disturbances
        \item $\pmb{\varepsilon}_M = [\varepsilon_{M,x}, \varepsilon_{M,y}, \varepsilon_{M,z}]^\top$: moment disturbances
        \item $\pmb{\varepsilon}_u = [\varepsilon_{u,1}, \varepsilon_{u,1}, \varepsilon_{u,3}, \varepsilon_{u,4}]$: action disturbances
    \end{itemize}
    \item \textbf{flight plan vector} $\mathbf{f}$ indicating the relative position between gates, detailed in \autoref{subsec:flightplan}.
\end{itemize}

The privileged observations are given by $\mathbf{o}_t^+ = [\mathbf{p_w}, \mathbf{p_g}, \mathbf{v_w}, \mathbf{v_g}, \pmb{\lambda}, \pmb{\Omega}, \pmb{\omega}, \mathbf{c}_e, \mathbf{d}]^T$, where it should be noted that $\pmb{\omega}$ and $\pmb{\Omega}$ represent the ground-truth rates, not the measured ones. The observations are $\mathbf{o}_t = [\mathbf{X}, \hat{\pmb{\Omega}}, \hat{\pmb{\omega}}]^T$, where $\mathbf{X} \in \{0,1\}^{H \times W}$ is a binary segmentation mask of the gates, generated by the segmentation model described in \autoref{subsec:segmentation_model}, $\hat{\pmb{\Omega}}$ are the body rates measured by the inertial measurement unit (IMU), and $\hat{\pmb{\omega}}$ are the propeller angular velocities measured by the electronic speed controller (ESC). We do not use the accelerometer, motivated by the observation of Bahnam \textit{et al.} \cite{aibeatshumandeWagter2025} that it tends to saturate on the used physical platform under high accelerations and vibrations. We define the information variable as $\mathbf{i}_t = \{\mathbf{o}_t, \mathbf{o}_t^+\} \setminus \{\mathbf{X}$\}, where the binary segmentation mask is excluded since its information is fully captured in $\mathbf{o}_t^+$, and thus still satisfies the conditional dependence of $\pmb{o}_t$ on $\pmb{s}_t$. Moreover, including it would significantly slow down training.

Lastly, the actions are given by $\mathbf{u}_t = [u_1, u_2, u_3, u_4]^\top$, which represent the normalized motor commands for each rotor, where $u_i \in [0, 1]$. These commands directly specify the desired motor power fraction, which the ESC regulates with pulse-width modulation (PWM), without any intermediate inner-loop controller, providing maximal control authority over the physical platform for high-speed, agile flight.

\subsection{Reward function}\label{subsec:reward_function}

In RL, the reward function is a key component, hand-designed to reflect the task objective. It encodes both the goals and constraints of the task, assigning positive rewards for desirable behavior and penalizing undesirable actions. Similar to Ferede \textit{et al.} \cite{ferede2025netrulealldomain}, our reward function consists of a progress reward, rate penalty and gate reward, and is given by:
\begin{align*}
r_t &= 5 \cdot r_{\text{prog}} - r_{\text{rate}} + 30 \cdot r_{\text{gate}} \label{eq:reward_main} \\
r_{\text{rate}} &= \frac{1}{2 \cdot f_c \cdot 10^5} \left(\exp\left(\min\left(\|\boldsymbol{\Omega}_t\|_1,\, 17\right)\right) - 1\right) \\
r_{\text{prog}} &= \|\mathbf{p}_{t-1,g}\|_2 - \|\mathbf{p}_{t,g}\|_2 \\
r_{\text{gate}} &=
\begin{cases}
1 - \dfrac{\max\left(|y_{t,g}|,\, |z_{t,g}|\right)}{d_g}, & \substack{\text{if gate} \\ \text{passed}} \\
0, & \substack{\text{otherwise}}
\end{cases}
\end{align*}
where $f_c=90\,\text{Hz}$ is the control frequency of the policy, and $r_{\text{rate}}$ is a rate penalty designed to discourage excessive angular velocities, which helps prevent both gyroscope saturation and excessive motion blur in the onboard camera. The term $r_{\text{gate}}$ provides an additional reward upon successfully passing through a gate, where $d_g$ is the effective size of the gate, with the reward being maximal at the center of the gate and decreasing linearly to zero toward its edges. 

To account for the thickness of the gate and the size of the drone, we introduce a \emph{pre-gate} and \emph{post-gate}, which are non-rendered virtual gates located at a specified distance before and after the gate, respectively. These gates provide the same reward $r_{\text{gate}}$ and collectively act as collision volumes with shaped rewards, spanning a total thickness $t_g$. 

The term $r_{\text{prog}}$ is a progress reward that encourages the drone to move toward the next pre-gate, serving as a dense shaping signal to accelerate training. Importantly, this term does not enforce a specific trajectory: any path toward the next gate results in the same accumulated return. For ease of implementation, the progress reward is set to zero between the pre- and post-gate. 

Crucially, unlike Geles \textit{et al.} \cite{geles2024demonstratingagileflightpixels} and Xing \textit{et al.} \cite{xing2024bootstrappingreinforcementlearningimitation} but similar to Romero \textit{et al.} \cite{romero2025dreamflymodelbasedreinforcement} and Krinner \textit{et al.} \cite{krinner2025acceleratingmodelbasedreinforcementlearning}, there is no explicit perception reward. We observe that the resulting policy naturally learns to orient its camera toward the gates to maximize gate visibility. Additionally, unlike Ferede \textit{et al.} \cite{ferede2025netrulealldomain}, we do not include a penalty for gate collisions. In our experiments, such penalties were found to hinder the learning process and provided no benefit compared to the natural truncation of the discounted return caused by an episode termination on collision.

Finally, an episode is terminated if any of the following conditions are met:
\begin{align*}
    \text{Gate collision:} & \quad
    \begin{cases}
\frac{\max\left(|y_{t,g}|,\, |z_{t,g}|\right)}{d_g} > 1, & \substack{\text{if gate} \\ \text{passed}} \\
\texttt{false}, & \substack{\text{otherwise}} \\
\end{cases} \\
\text{Ground collision:} \quad & z_w > -0.5 \;\land\; \big( v_{z,w} > 1.0 \\
& \lor\; |\phi_w| > \frac{\pi}{3} \;\lor\; |\theta_w| > \frac{\pi}{3} \big)
\end{align*}
where $z_w = 0$ denotes the ground plane. Episodes terminating due to either condition result in a final reward of zero. The gate collision condition also applies to pre- and post-gates. The ground collision condition is shaped to discourage fast flight with high pitch or roll near the ground. This is discouraged as the drone starts from a raised podium in the real world, and such contact forces are not simulated. Using this shaped collision condition, the drone learns to ascend first before accelerating forward, rather than immediately flying toward the gate.

\subsection{Quadcopter dynamics}\label{subsec:dynamics}
Since the policy is trained in simulation, accurate modeling of the quadcopter dynamics is crucial, particularly because motor commands are executed directly. Our dynamics model largely builds on the frameworks from Bahnam \textit{et al.} \cite{aibeatshumandeWagter2025} and Ferede \textit{et al.} \cite{ferede2025netrulealldomain}; for full details, we refer the reader to those sources. Following the approach of Bahnam \textit{et al.} \cite{aibeatshumandeWagter2025}, we represent the rotation described by $\pmb{\lambda}$ using quaternions $\mathbf{q} = [q_w, q_x, q_y, q_z]^T$ and incorporate gyroscopic effects. Furthermore, we introduce additional disturbances $\pmb{\varepsilon}$:
\begin{align*}
    \dot{\mathbf{x}}_w &= \mathbf{v}_w \\
    \dot{\mathbf{q}} &= \frac{1}{2} \mathbf{q} \otimes [0\; p\; q\; r]^T \\
    \dot{\mathbf{v}}_i &= g \mathbf{e}_3 + \mathbf{R}(\pmb{\lambda}) \mathbf{F} + \pmb{\varepsilon}_a \\
    \dot{\pmb{\Omega}} &= \mathbf{M} + \pmb{\varepsilon}_M \\
    \omega_i &= \frac{\omega_{ci} - \omega_i}{\tau}
\end{align*}

where $\mathbf{R}$ is the rotation matrix and $g$ is the gravitational acceleration constant. The forces $\mathbf{F}$, moments $\mathbf{M}$ and steady motor response $w_{ci}$ are given by:

\begin{align*}
    \mathbf{F} = \begin{bmatrix}
    -k_x v^B_x \sum_{i=1}^4 \omega_i - k_{x2} v^B_x |v^B_x| \\
    -k_y v^B_y \sum_{i=1}^4 \omega_i - k_{y2} v^B_y |v^B_y| \\
    -k_\omega \left(1 + k_{\alpha} \alpha + k_{\text{hor}} \mu_{xx,yy} \right) \sum_{i=1}^4 \omega_i^2 - k_{v2} v^B_z |v^B_z|
    \end{bmatrix} \\
\end{align*}
\begin{align*}
    \text{with}\quad\quad\quad\quad\quad\quad\quad\quad\quad\\
    \alpha = \tan^{-1}\Big(\frac{v^B_z}{r \bar{\omega}}\Big) \quad \quad\quad\quad\quad\quad \\ 
    \mu_{xx,yy} = \tan^{-1}\Big(\frac{v^{B2}_x+v^{B2}_y}{r \bar{\omega}}\Big) \quad \text{with} \quad \bar{\omega}=\sum_{i=1}^4 \omega_i
\end{align*}

\begin{align*}
    \pmb{M} = \begin{bmatrix}
-k_{p1} \omega_1^2 - k_{p2} \omega_2^2 + k_{p3} \omega_3^2 + k_{p4} \omega_4^2 + J_x q r \\
-k_{q1} \omega_1^2 + k_{q2} \omega_2^2 - k_{q3} \omega_3^2 + k_{q4} \omega_4^2 + J_y p r \\
-k_{r1} \omega_1 + k_{r2} \omega_2 + k_{r3} \omega_3 - k_{r4} \omega_4 - \\ 
k_{r5} \dot{\omega}_1 + k_{r6} \dot{\omega}_2 + k_{r7} \dot{\omega}_3 - k_{r8} \dot{\omega}_4 + J_z p q
\end{bmatrix}
\end{align*}

\begin{align*}
    \omega_{ci}=(\omega_{\text{max}}-\omega_{\text{min}})\sqrt{k \tilde{u_i}^2 + \left(1-k_l\right)\tilde{u_i}} + \omega_{\text{min}} \\
    \text{with}\quad\quad\quad\quad\quad\quad\quad\quad\quad \\
    \tilde{u_i} = \min\left(\max\left(u_i + \varepsilon_{u,i}, 0\right), 1\right) \quad\quad\quad
\end{align*}
where $v_{x,b}$, $v_{y,b}$ and $v_{z,b}$ are the velocity components in body frame.

The values of the parameters used for both policy training and simulation experiments are given in \autoref{tab:drone_params}. These parameters together make up the previously mentioned dynamics parameters $\mathbf{d} = [k_w, k_x, k_y, \dots, \omega_{\text{max}}, k, \tau]^T$. The simulation is implemented using JAX in Python, and is discretized using fourth-order Runge–Kutta integration with a timestep of $2.2 \, \text{ms}$.

\begin{table}[ht]
\centering
\caption{Quadcopter dynamical parameters used for both policy training and simulation experiments.}
\label{tab:drone_params}
\begin{tabular}{ll ll}
\toprule
\textbf{Param} & \textbf{Value} & \textbf{Param} & \textbf{Value} \\
\midrule
$k_w$             & $1.55 \times 10^{-6}$   & $k_{x2}$         & $4.10 \times 10^{-3}$ \\
$k_x$, $k_y$      & $5.37 \times 10^{-5}$   & $k_{y2}$         & $1.51 \times 10^{-2}$ \\
$k_\text{angle}$  & $3.145$                 & $k_\text{hor}$   & $7.245$ \\
$k_{v2}$          & $0.00$                  & $J_x$            & $-0.89$ \\
$J_y$             & $0.96$                  & $J_z$            & $-0.34$ \\
$\omega_{\min}$   & $341.75$                & $\omega_{\max}$  & $3100.00$ \\
$k$               & $0.50$                  & $\tau$           & $0.03$ \\
$k_{p1}$          & $4.99 \times 10^{-5}$   & $k_{p2}$         & $3.78 \times 10^{-5}$ \\
$k_{p3}$          & $4.82 \times 10^{-5}$   & $k_{p4}$         & $3.83 \times 10^{-5}$ \\
$k_{q1}$          & $2.05 \times 10^{-5}$   & $k_{q2}$         & $2.46 \times 10^{-5}$ \\
$k_{q3}$          & $2.02 \times 10^{-5}$   & $k_{q4}$         & $2.57 \times 10^{-5}$ \\
$k_{r1}$--$k_{r4}$ & $3.38 \times 10^{-3}$   & $k_{r5}$--$k_{r8}$ & $3.24 \times 10^{-4}$ \\
\bottomrule
\end{tabular}
\end{table}

\subsection{Segmentation masks}\label{subsec:segmentation_model}

As pixel level input, SkyDreamer receives a segmentation mask $\mathbf{X}$. Unlike the extrinsic parameters, we choose to calibrate the intrinsic parameters of the camera to maintain a consistent camera matrix for SkyDreamer’s inputs, as they are straightforward to obtain and remain constant over time. Using these calibrated intrinsics, the RGB images are undistorted, cropped, and mapped to a fixed-resolution image of size $H \times W$, such that the resulting image satisfies a nominal pinhole camera model with intrinsic matrix:
\[
\mathbf{K} = 
\begin{bmatrix}
f_x & 0 & c_x \\
0 & f_y & c_y \\
0 & 0 & 1
\end{bmatrix}
=
\begin{bmatrix}
\frac{25}{64}W & 0 & 0.5W \\
0 & \frac{25}{64}H & 0.5H \\
0 & 0 & 1
\end{bmatrix},
\]
where $f_x$ and $f_y$ are the focal lengths in pixels and $c_x$, $c_y$ are the principal point coordinates in pixels. The factor $\frac{25}{64}$ is empirically chosen such that an image of $64 \times 64$ pixels corresponds to a focal length of $25$, providing a balanced field of view that maximizes scene coverage while avoiding black borders in the undistorted images. By mapping all images to a shared nominal intrinsic matrix and randomizing extrinsic parameters during training, SkyDreamer ensures robustness to extrinsic variations and invariance to intrinsic parameters through calibration, enabling generalization of the same model across different drones.

Subsequently, the fixed-resolution image is passed to a segmentation model called GateNet that produces a binary segmentation mask $\mathbf{X} \in \{0,1\}^{H \times W}$, indicating which pixels correspond to gates. GateNet follows a U-Net architecture, similar to the designs used in De Wagter \textit{et al.} \cite{dewagteralphapilot2019} and Bahnam \textit{et al.} \cite{aibeatshumandeWagter2025} of which the implementation details are provided in \aref{app:gatenet}. The resulting masks are resized to a resolution of $64 \times 64$ to speed up the training process of SkyDreamer.

\subsection{Visual augmentations}
During training, segmentation masks are rendered in batches using PyTorch3D. However, these masks are idealized and differ significantly from real-world masks, causing a large visual sim-to-real gap. To bridge this gap, we employ CycleGAN \cite{cyclegan}, a generative adversarial network (GAN) that learns to translate images between two unpaired domains, in our case, from synthetic segmentation masks to more realistic masks and vice versa, without requiring paired real and rendered masks. 

To further improve realism, we use a stochastic CycleGAN (StochGAN) \cite{stochastic_gan}, an extension of CycleGAN that adds noise to the inputs. This stochasticity enhances the variation and realism of generated masks, better approximating real-world appearance and variance. Although Almahairi \textit{et al.} \cite{stochastic_gan} also proposes a more advanced AugmentedGAN, we opted for StochGAN due to its simpler implementation and effective performance. The GAN is trained on approximately 4000 unpaired real and rendered images, which require no manual labeling and are therefore inexpensive to collect. Further implementation details are provided in \aref{app:stochgan}.

We observed that real-life masks were often thinner, a detail not captured well by the GAN alone. To address this, we randomly erode the masks by $1$ pixel -- using average pooling followed by thresholding -- $50 \%$ of the time at a mask resolution of $64 \times 64$, introducing variation in the mask thickness. Since the erosion level typically remains stable over short periods, we hold it fixed for an average of $100$ environment steps.

Lastly, to simulate the rolling shutter effect of the onboard camera, we account for two dominant effects: a horizontal shear induced by the yaw rate in camera frame, and a vertical scaling induced by the pitch rate. These arise from the line-by-line capture of a rolling shutter camera under rotational motion. To model these effects, we introduce a rolling shutter parameter $s \in [0,0.02]$, which also aims to capture additional disturbances associated with rolling shutter through randomization.

We formalize our approximation of the rolling shutter effect with an affine transformation given by the following matrix:
\begin{align*}
\mathbf{A} = 
\begin{bmatrix}
1 & -s \, r_c & \tfrac{W}{2}\, s \, r_c \\
0 & 1 + s \, q_c & -\tfrac{H}{2} \, s \, q_c
\end{bmatrix},
\end{align*}
where $r_c$ and $q_c$ denote the yaw and pitch rates in the camera frame, respectively. This transformation is then applied to the rendered segmentation mask:
\begin{align*}
\mathbf{X}' = \operatorname{warp\_affine}(\mathbf{X}, \mathbf{A})
\end{align*}
which produces a horizontally sheared and vertically scaled segmentation mask.

\subsection{SkyDreamer architecture}

\begin{figure*}[t]
    \centering
    \begin{subfigure}[t]{0.48\textwidth}
        \centering
        \includegraphics[width=\textwidth]{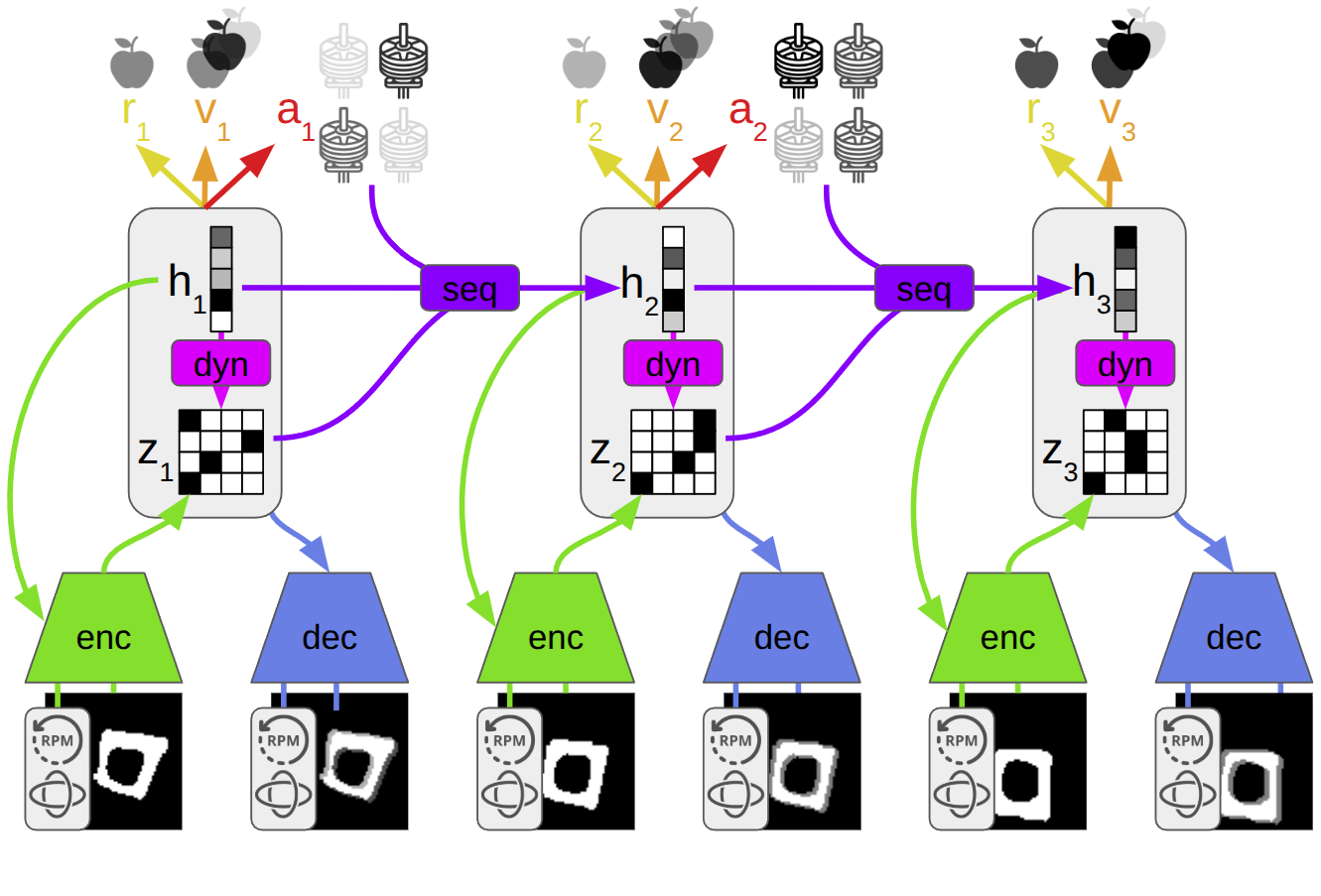}
        \caption{Standard DreamerV3 architecture for ADR. In standard DreamerV3, the world model is not guided by privileged observations $\mathbf{o}_t^+$.}
        \label{fig:standard_dreamerv3_adr}
    \end{subfigure}
    \hfill
    \begin{subfigure}[t]{0.48\textwidth}
        \centering
        \includegraphics[width=\textwidth]{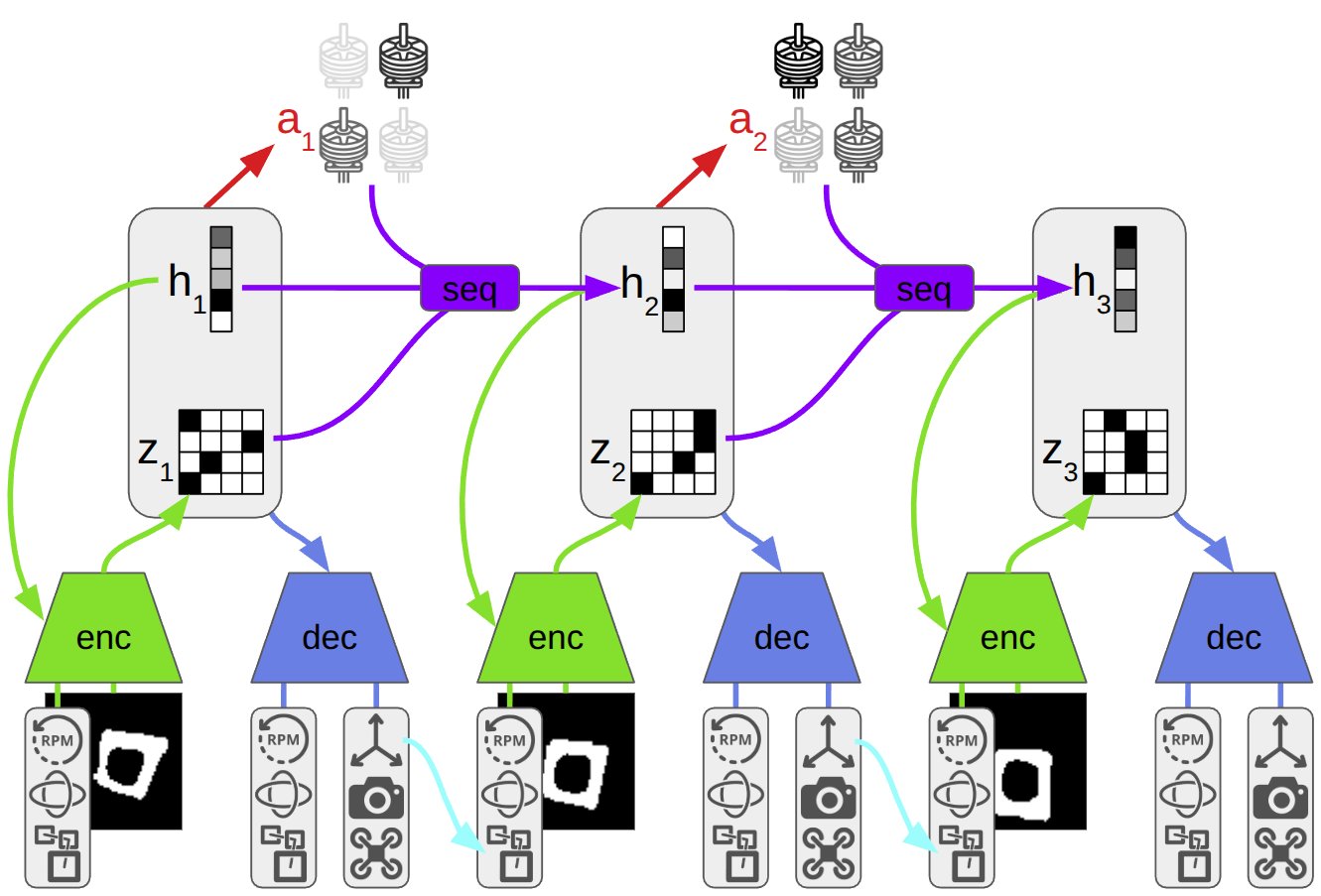}
        \caption{SkyDreamer at deployment. The dynamics predictor is not used, as observations are always available. The flight plan vector is updated based on the decoded states as indicated by the cyan arrow.}
        \label{fig:skydreamer_deployment}
    \end{subfigure}
    \vspace{0.5em}
    \begin{subfigure}[t]{0.48\textwidth}
        \centering
        \includegraphics[width=\textwidth]{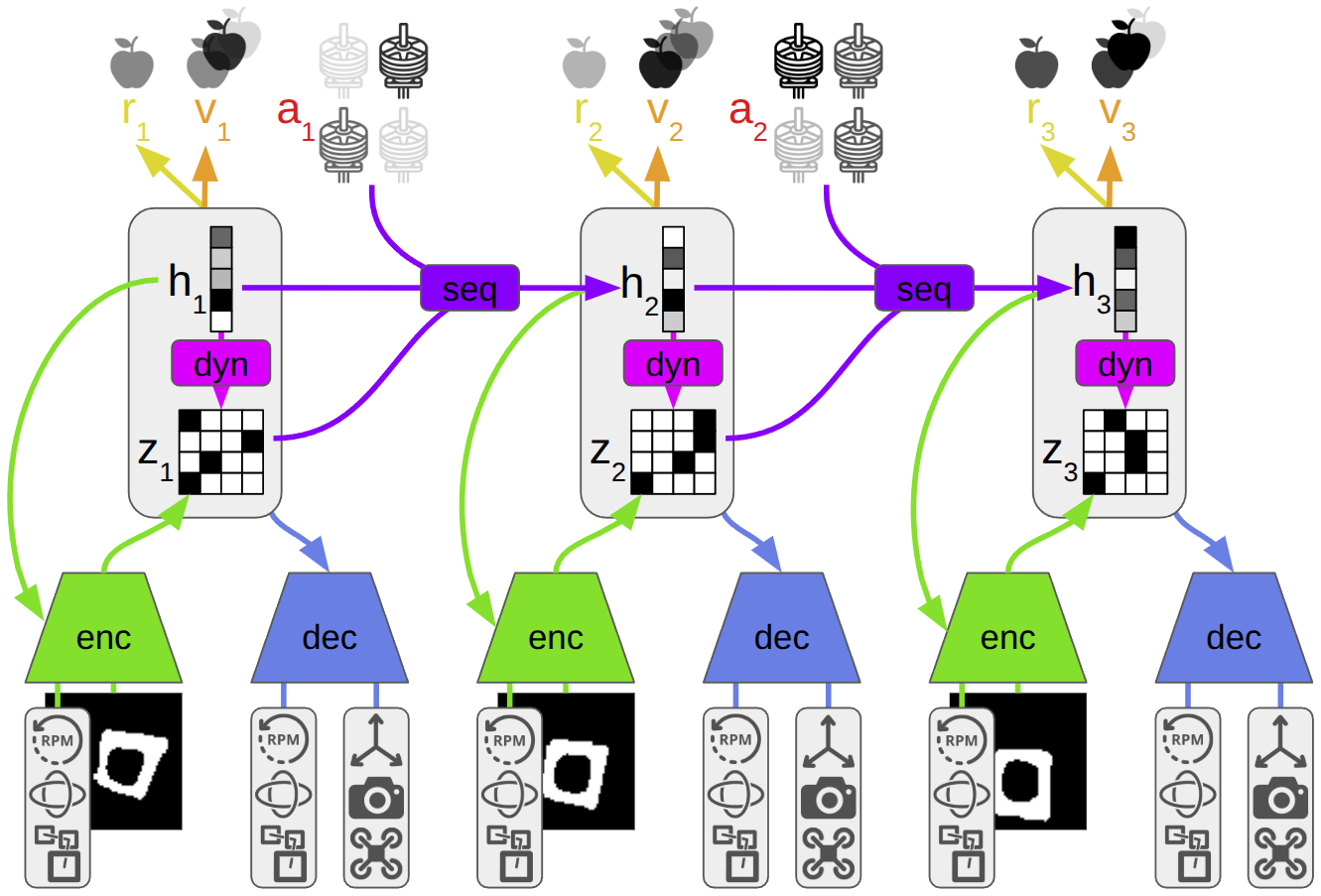}
        \caption{SkyDreamer during world model learning. The dynamics predictor is trained to match the encoded inputs. Additionally, the sequence model, encoder, decoder, reward predictor, and continue predictor are trained.}
        \label{fig:skydreamer_wm_learning}
    \end{subfigure}
    \hfill
    \begin{subfigure}[t]{0.48\textwidth}
        \centering
        \includegraphics[width=\textwidth]{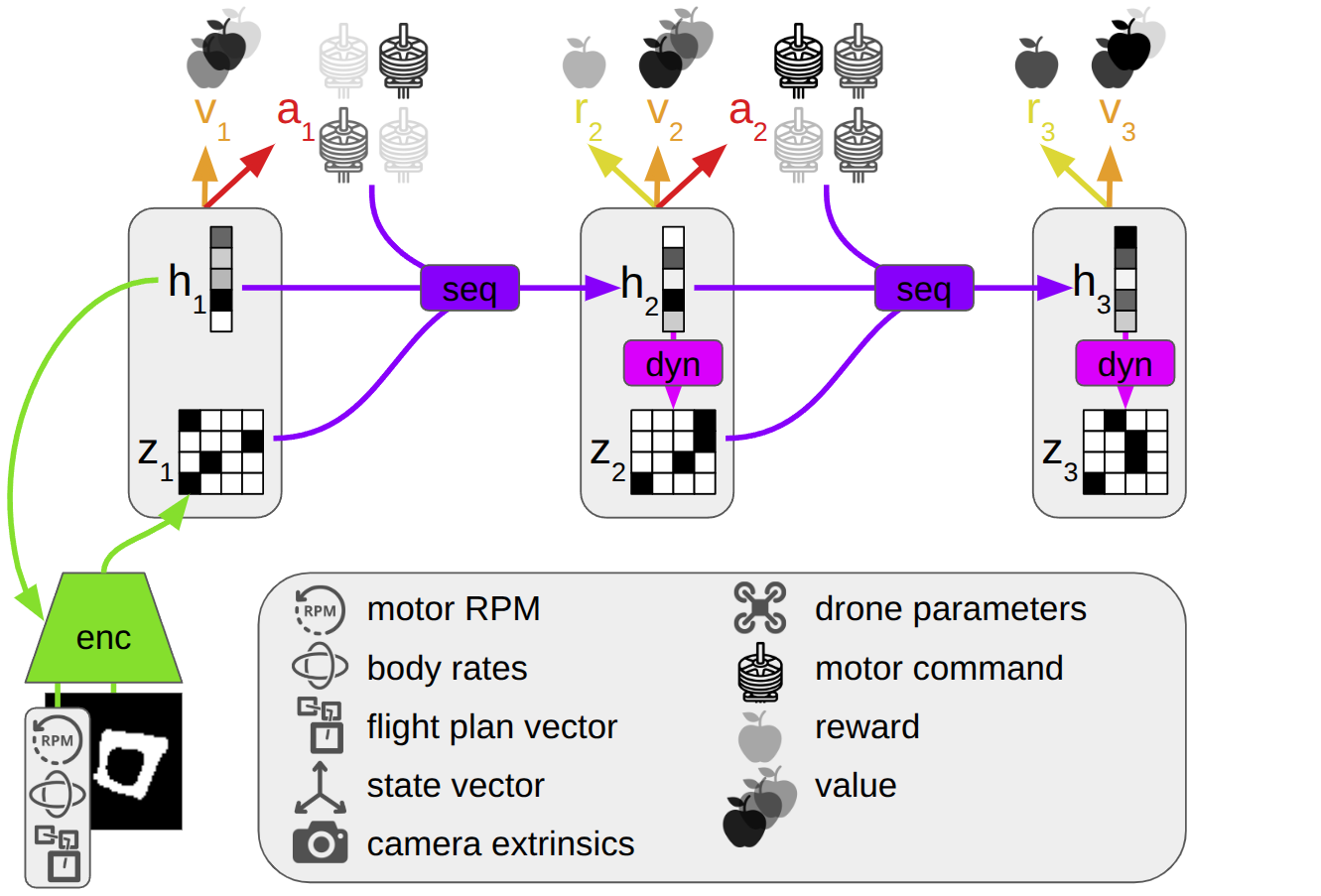}
        \caption{SkyDreamer during actor-critic learning. The dynamics predictor replaces the encoder to generate imagined rollouts.}
        \label{fig:skydreamer_ac_learning}
    \end{subfigure}
    \caption{Comparison of standard DreamerV3 and SkyDreamer across different stages. The world model encodes segmentation masks and measurements such as motor RPMs, body rates, and the flight plan vector into discrete representations $z_t$. The next hidden state $h_{t+1}$ is predicted from the previous hidden state $h_t$, action $a_t$ and discrete representation $z_t$ using the sequence model. The inputs $\hat{x}_t$ are reconstructed to guide the hidden representations. In SkyDreamer, the privileged observations, including ground-truth state, camera and dynamical parameters, are decoded as $\mathbf{o}_t^+$. During actor-critic learning, imagined rollouts are generated solely using the dynamics predictor without feedback from the environment. Images inspired by Hafner \textit{et al.} \cite{dreamerv3_nature}.}
    \label{fig:skydreamer_overview}
\end{figure*}

SkyDreamer's architecture is almost entirely based on Informed Dreamer \cite{lambrechts2024informedpomdpleveragingadditional}, which is largely based on DreamerV3 \cite{dreamerv3_nature}. In this section, we provide a high level overview and intuitive explanation of the architecture most relevant to our work. For detailed information on the architecture, policy learning and loss functions, we refer the reader to Hafner \textit{et al.} \cite{dreamerv3_nature} and Lambrechts \textit{et al.} \cite{lambrechts2024informedpomdpleveragingadditional}. 

The core component of DreamerV3 is its world model, which encodes observations into a latent representation and predicts future latent states based on actions, both with and without receiving further observations. The training process continuously alternates between three main stages: (1) data collection, (2) world model learning, and (3) actor-critic learning, which are illustrated in \autoref{fig:skydreamer_overview}. In the data collection phase, the current policy interacts with the environment to collect rollouts, which are stored in a replay buffer, making DreamerV3 an off-policy algorithm. During world model learning, batches of trajectories are uniformly sampled from the replay buffer to train the world model. Finally, in the actor-critic phase, learning occurs entirely through latent imagination: the world model simulates rollouts without environment feedback, enabling sample efficient model-based RL. All components of SkyDreamer are implemented as multi-layer perceptrons (MLPs), unless otherwise noted.

The key element of the world model in DreamerV3 is a \textbf{recurrent state-space model} (RSSM), which consists of the following components:

\begin{align*}
    \text{Encoder:} &\quad \mathbf{z_t} \sim q^e_\theta(\cdot | \mathbf{h}_t, \mathbf{o}_t) \\
    \text{Sequence model:} &\quad \mathbf{h_t} = f^e_\theta(\mathbf{h}_{t-1}, \mathbf{z}_
    {t-1}, \mathbf{u}_{t-1}) \\
    \text{Dynamics predictor:} &\quad \hat{\mathbf{z}_t} \sim p_\theta^d(\cdot | \mathbf{h_t}) \\
\end{align*}

Here, $q^e_\theta$ is the \textbf{encoder}, which encodes the observation $\mathbf{o}_t$ along with the hidden state $\mathbf{h}_t$ into a discrete stochastic latent representation $\mathbf{z}_t$. This structure resembles a vector-quantized variational autoencoder (VQ-VAE) \cite{vqvae}. For image observations, it uses a convolutional neural network (CNN) rather than an MLP. The function $f^e_\theta$ is the \textbf{sequence model}, implemented as a single-layer gated recurrent unit (GRU) \cite{gru}, which predicts the next hidden state $\mathbf{h}_t$ given the previous hidden state $\mathbf{h}_{t-1}$, previous latent state $\mathbf{z}_{t-1}$, and previous action $\mathbf{u}_{t-1}$. Finally, $p^d_\theta$ is the \textbf{dynamics predictor}, which predicts an estimate of the latent representation $\hat{\mathbf{z}}_t$ based solely on the hidden state $\mathbf{h}_t$. This component is crucial, as it allows the model to roll out future trajectories in latent space without further observations from the environment, which is often referred to as \textit{latent imagination} or \textit{dreaming} and is illustrated in \autoref{fig:skydreamer_ac_learning}.

The RSSM also employs a \textbf{decoder} to guide the learning process. Crucially, following informed Dreamer \cite{lambrechts2024informedpomdpleveragingadditional}, SkyDreamer decodes to the privileged information $\mathbf{i}_t$ rather than only the raw observations $\mathbf{o}_{t}$:
\begin{align*}
    \text{Decoder:} \quad & \mathbf{i}_t \sim p_t^i(\cdot | \mathbf{h_t}, \mathbf{z_t})
\end{align*}
By decoding to this privileged information, the world model is encouraged to extract more meaningful and task-relevant features from the inputs. This key difference from standard DreamerV3 is visually evident when comparing \autoref{fig:standard_dreamerv3_adr} and \autoref{fig:skydreamer_wm_learning}. In addition to improving learning speed, final performance and adaptability to drone-specific parameters, this also results in a more interpretable world model. Whereas decoding to images often leads to blurry reconstructions, decoding to states such as position, velocity, and attitude allows a direct inspection of what the world model believes the current state to be. 

Additionally, this interpretability offers a significant practical advantage during debugging: state estimation related errors can be identified and resolved without additional real-world flights. Instead, one can replay with an updated world model using the same sequence of observations and actions to verify potential improvements. This greatly accelerates the iteration cycle when addressing vision-related sim-to-real transfer issues, a common challenge in end-to-end vision-based learning approaches.

To support actor-critic learning during latent imagination, the world model is also trained to predict the reward and whether a state is terminal. This is achieved using a \textbf{reward predictor} $p_\theta^r$ and a \textbf{continue predictor} $p_\theta^c$:
\begin{align*}
    \text{Reward predictor: } \quad & \hat{r_t} \sim p_\theta^r(\cdot | \mathbf{h_t}, \mathbf{z_t}) \\
    \text{Continue predictor: } \quad & \hat{c_t} \sim p_\theta^c(\cdot | \mathbf{h_t}, \mathbf{z_t})
\end{align*}
where $p_\theta^c$ predicts the continuation probability $\hat{c_t} \in [0,1]$, indicating whether the episode is expected to continue.

During actor-critic learning, the \textbf{actor} is optimized using the Reinforce estimator \cite{reinforce}. Actor-critic learning is performed entirely in latent space through imagined rollouts, as illustrated in \autoref{fig:skydreamer_ac_learning}. These rollouts begin from hidden states obtained during world model training and are propagated forward for $16$ steps ($0.18\,\text{s}$) using the dynamics predictor, without any feedback from the environment.

The actor and critic are formalized as:
\begin{align*}
    \text{Actor: } \quad & \mathbf{u}_t \sim \pi_\theta(\cdot | \mathbf{h}_t,  \mathbf{z}_t) \\
    \text{Critic: } \quad & v_\psi(V_t|\mathbf{h}_t,  \mathbf{z}_t) \\
\end{align*}

where the actor is implemented as a Gaussian policy with mean $\pmb{\mu}_t = \mathbb{E}_{\pi_\theta} \left[ \mathbf{u}_t | \mathbf{h}_t,  \mathbf{z}_t \right]$ and variance $\pmb{\sigma}_t^2 =\mathbb{E}_{\pi_\theta} \left[ \left( \mathbf{u}_t - \pmb{\mu}_t \right)^2 \mid \mathbf{h}_t, \mathbf{z}_t \right]$, and the critic estimates the value of the current state, i.e. the expected discounted return $V_t=\sum_{\tau=0}^\infty\gamma^\tau r_{t+\tau}$, with $\gamma=0.997$ the discount factor.

Although the standard implementation produced performant policies in simulation, we observed that they often converged to near bang-bang behavior, where the motors are either fully activated or deactivated. While such policies can be optimal in simulation, in the real world they lead to overheating or even burning the motors.

To mitigate this, we introduce a smoothness regularization loss inspired by Mysore \textit{et al.} \cite{smooth_policies}, encouraging smoother changes in control outputs over time. However, DreamerV3 includes an entropy loss that encourages high-variance (i.e., exploratory) policies to discover new strategies. Naively applying a smoothness penalty to the full policy distribution would therefore conflict with this entropy objective.

To avoid this conflict, we apply the smoothness loss $\mathcal{L}_{smooth}$ only to the mean of the policy and add $\mathcal{L}_{smooth}$ to the standard policy loss of DreamerV3:
\begin{align*}
    \mathcal{L}_{smooth} = \lambda_{smooth} \, \mathbb{E}_t \left[ \left| \left| \pmb{\mu}_t  - \pmb{\mu}_{t-1}  \right| \right|^2_2 \right]
\end{align*}
where $\lambda_{smooth}=0.002$ determines the weighting of the regularization. This regularization is applied to the imagined rollouts used during actor-critic learning. At inference time and during evaluation, we use the deterministic version of the policy ($\pmb{\sigma}_t=0$) to produce even smoother control.

\subsection{Flight Plan Logic}\label{subsec:flightplan}

Another key limitation of prior end-to-end vision-based ADR approaches is their inability to resolve visual ambiguities: situations where different parts of the track produce similar visual observations. To address this, we introduce a \textit{flight plan logic} that provides the policy with structured information about its progress along the track, inspired by \cite{robinrlfirstrlone, ferede2025netrulealldomain}.

Specifically, the policy is augmented with a flight plan vector $\mathbf{f}$ encoding the relative positions and yaw angles of upcoming gates, without containing the drone’s current position, attitude, or any related state variables. For each of the next three gates, the input includes the difference in 3D position and yaw between gate $i$ and gate $i+1$, along with the absolute position and yaw of gates $i$ through $i+2$ in the world frame:
\begin{align*}
\mathbf{f}_i = \bigl[
\mathbf{p}_{i}^g - \mathbf{p}_{i-1}^g,\
\pmb{\psi}_{i}^g - \pmb{\psi}_{i-1}^g,\
\dots,\
\pmb{\psi}_{i+2}^g - \pmb{\psi}_{i+1}^g,\\
\mathbf{p}_{i}^g,\
\pmb{\psi}_{i}^g,\
\dots,\
\pmb{\psi}_{i+2}^g
\bigr]^T
\end{align*}
where $\mathbf{p}_i^g$ denotes the position and $\pmb{\psi}_i^g$ the yaw angle of gate $i$ in the world frame. This flight plan vector $\mathbf{f}_i$ is provided as input to SkyDreamer alongside the other observations $\mathbf{o}_t$.

To determine when a gate has been passed and update the flight plan vector accordingly, we inspect the gate-relative position $\hat{\mathbf{x}}_g$ estimated by SkyDreamer. Once the drone has passed through the current gate, the flight plan index $i$ is incremented, which is empirically defined as:
\begin{align*}
    \hat{\mathbf{x}}_g > -0.15 \, m
\end{align*}
since the nominal threshold of  $\hat{\mathbf{x}}_g > 0.0 \, m$ occasionally failed to trigger due to small state estimation errors in SkyDreamer, as observed in one real-world flight. During training, however, we do not rely on SkyDreamer’s decoded states, which are often too inaccurate in the early part of training. Instead, the flight plan index $i$ is incremented randomly between the pre- and post-gate, simulating that the index does not always increment precisely when passing the gate.

This proposed flight plan logic not only helps resolve visual ambiguities, but also potentially enables generalization to arbitrary tracks. Moreover, it facilitates decoding to a global state in world frame under visual ambiguity or when training a single model across multiple tracks. We argue that such structured integration of track geometry is essential for developing end-to-end vision-based ADR systems capable of flying arbitrary unseen tracks with approximately known gate locations.

\section{Experiments and Results}

\begin{figure*}[ht]
    \centering
    \begin{subfigure}[t]{0.67\linewidth}
        \centering
        \includegraphics[width=\linewidth]{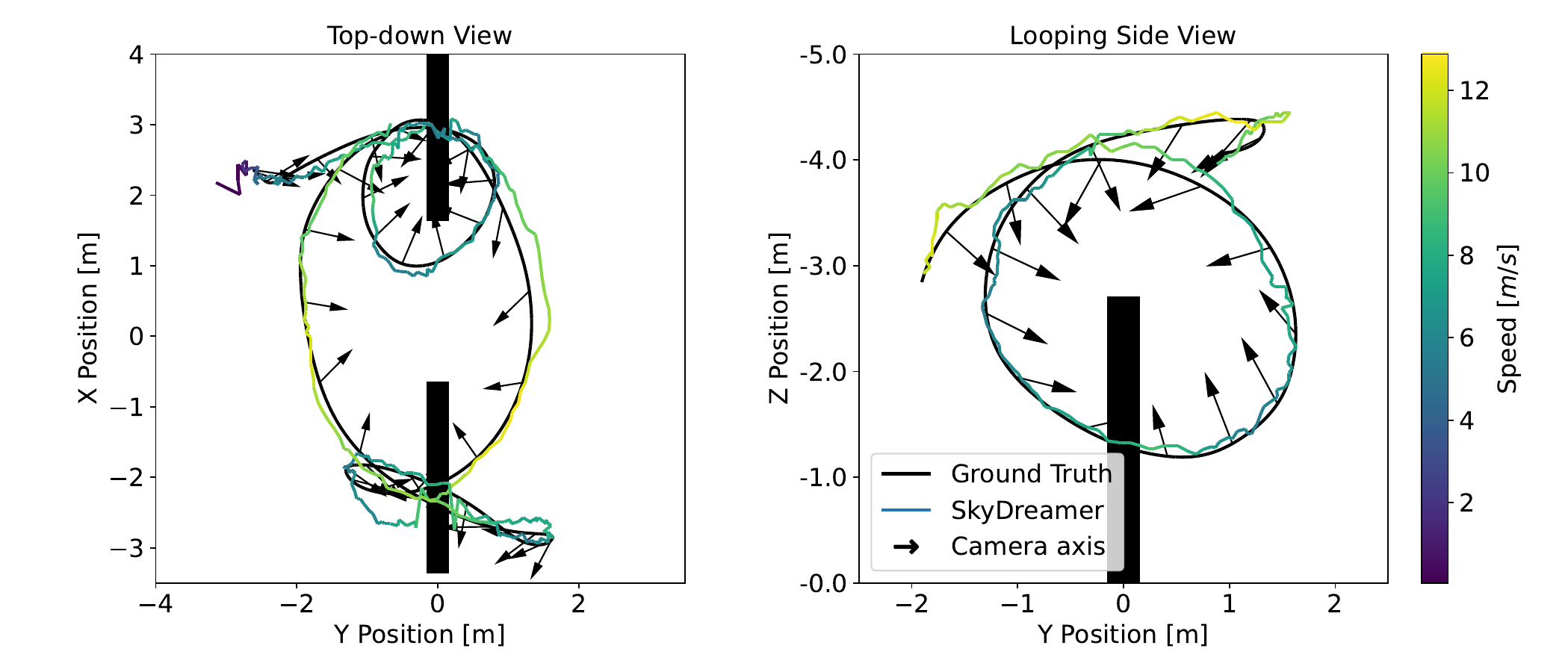}
    \end{subfigure}
    \hfill
    \begin{subfigure}[t]{0.32\linewidth}
        \centering
        \includegraphics[width=\linewidth,trim={5cm 0 0 0},clip]{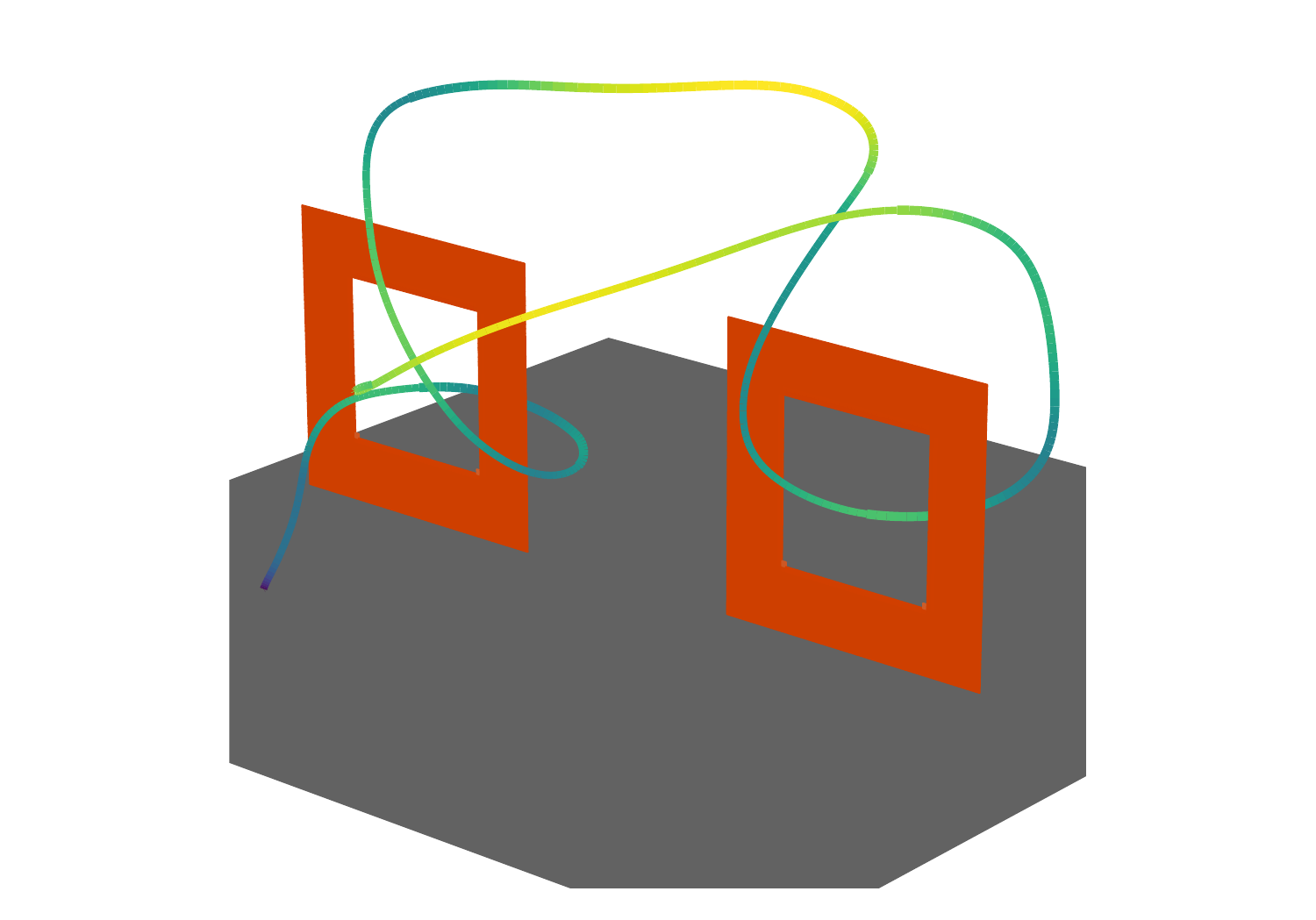}
    \end{subfigure}
    \caption{Simulated ladder inverted loop track flown by SkyDreamer. The flight begins by passing through the first gate, followed by a tight ladder maneuver and flying over it. SkyDreamer then flies over the second gate, performs a maneuver similar to a split-S through it, and flies back over the gate, completing the inverted loop. The lap is completed by passing through the first gate again. In the left image, the black lines indicate ground-truth trajectory, while the colored lines show SkyDreamer's position and velocity estimates, with color denoting speed. The black blocks mark the gate locations with exaggerated thickness. The black arrows indicate the camera principal axis direction, which show that SkyDreamer naturally orients its camera toward the gates: an emergent behavior achieved without any explicit perception reward. The right image presents a 3D render of one lap flown by SkyDreamer, where the colored line represents the ground-truth trajectory and corresponding speed.}
    \label{fig:perception_ladder_sim_combined}
\end{figure*}

We extensively evaluate SkyDreamer on two tracks and with two gate types, in simulation and in the real world: the \textit{ladder inverted loop} track and the \textit{inverted loop} track, visualized in \autoref{fig:perception_ladder_sim_combined} and \autoref{fig:inverted_looping_real_combined_orange}. During training, additional invisible gates are added to enforce the correct flight maneuvers. Both tracks highlight a common challenging drone racing maneuver at the second gate: the split-S. Within our limited $8 \times 8 \text{m}$ testing space, we extend this split-S into a full inverted loop to create a circular track. At the first gate, the ladder track emphasizes precise control close to a gate, while the inverted loop track tests high-speed flight through a gate. Finally, we test SkyDreamer on the \textit{big track}, shown in \autoref{fig:perception_uvv_combined}, demonstrating its ability to handle more complex layouts and higher-speed flight.

Importantly, these tracks would be difficult -- or even impossible -- to master for end-to-end vision-based approaches without our proposed flight plan logic: the drone encounters visually identical observations at different points along the track but require taking different flight paths, which would otherwise introduce ambiguity. By leveraging this flight plan logic, SkyDreamer learns to resolve such ambiguity in both simulation and the real world without external aid.

\subsection{Experimental setup}

For our experiments, we train SkyDreamer starting from the latest JAX implementation of DreamerV3\footnote{ \url{https://github.com/danijar/dreamerv3/commit/cdf570902b1eaba193cc8ef69426cd4edde1b0bc}} for a total of $17$ million environment steps, using the standard \texttt{size12m} parameter model and default hyperparameters unless otherwise noted. We set \texttt{replay\_context} to $16$, \texttt{train\_ratio} to $128$, \texttt{slowtar} to \texttt{true}, and use a replay buffer size of $10 \cdot 10^6$ representing $31 \, \text{h}$ of flight time. 

We use environment settings and initial states as specified in \autoref{tab:randomization_ranges}. During training, $70\%$ of initial states are sampled from the training column, where initialization can occur in front of any gate, and $30\%$ from the evaluation column, where initialization always occurs at the start gate.

\begin{table}[ht!]
\centering
\footnotesize
\caption{Randomization, parameter and initial states. "Train" refers to the settings used for collecting data that is stored in the replay buffer during training. "Evaluation" refers to the settings used during evaluation and simulation experiments. For disturbances where a frequency is specified, 90 Hz indicates the value changes every timestep, whereas 1 Hz indicates a 1/100 probability of change per timestep.}\label{tab:randomization_ranges}
\begin{tabular}{lccr}
\toprule
\textbf{Parameter} & \textbf{Train} & \textbf{Evaluation} & \textbf{Unit} \\
\midrule
$\phi_c$, $\psi_c$       & $[-5, 5]$       & $[-5, 5]$       & deg \\
$\theta_c$               & $[45, 55]$      & $[45, 55]$      & deg \\
$\omega_\text{min}$, $\omega_\text{max}$ & $\pm 20\%$     & $\pm 20\%$     & \% rand \\
$\mathbf{d} \setminus \{\omega_\text{min}, \omega_\text{max}\}$                & $\pm 30\%$      & $\pm 20\%$      & \% rand \\
$\pmb{\varepsilon}_a\text{ (1Hz)}$    & $[-3,3]$  & $[-2,2]$  & m/s\textsuperscript{2} \\
$\pmb{\varepsilon}_M\text{ (1Hz)}$    & $[-3,3]$  & $[-2,2]$  & rad/s\textsuperscript{2} \\
$\pmb{\varepsilon}_M\text{ (90Hz)}$    & $[-125,125]$  & $[-100,100]$  & rad/s\textsuperscript{2} \\
$\pmb{\varepsilon}_u\text{ (90Hz)}$    & $[-0.2,0.2]$  & $[-0.2,0.2]$  & - \\
$d_g$                    & $0.8$           & $1.0$           & m \\
$t_g$                    & $0.8$           & $0.8$           & m \\
\midrule
\textbf{Initial state} &  &  &   \\
\midrule
$x_g$       & $[-4.0, -2.0]$       & $[-4.0, -2.0]$       & m \\
$y_g$       & $[-1.0, 1.0]$       & $[-1.0, 1.0]$       & m \\
$z_g$       & $[0.0, 1.3]$       & $[0.7, 1.3]$       & m \\
$v_{x,w}$, $v_{y,w}$, $v_{z,w}$       & 0.0 & 0.0 & m/s \\
$p$, $q$, $r$       & $[-0.1, 0.1]$       & $[-0.1, 0.1]$       & rad/s \\
$w_i$       & $[0.25, 0.5]$       & $[0.25, 0.5]$       & $\omega_{\text{max}}$ \\
$\phi_w$, $\theta_w$, $\psi_g$  & $[-\pi/9, \pi/9]$ & $[-\pi/9, \pi/9]$ & rad \\
\bottomrule
\end{tabular}
\end{table}

Training is conducted in three phases on a 40GB partition with 56 Shared Multiprocessors on an NVIDIA A100 80GB GPU for roughly $50$ hours. In the initial warm-up phase, we follow the default DreamerV3 settings. After $8$ million steps, we increase the batch length for world model training from 64 to 256 to better capture long-term dependencies, which is especially useful for parameter identification. After 13 million steps, we reduce the entropy coefficient from $3 \cdot 10^{-4}$ to $1 \cdot 10^{-5}$ and the learning rate from $4 \cdot 10^{-5}$ to $2 \cdot 10^{-6}$ to stabilize and fine-tune the policy for smoother and more fine-grained control.

\begin{figure*}[ht!]
    \centering
    \includegraphics[width=0.99\linewidth]{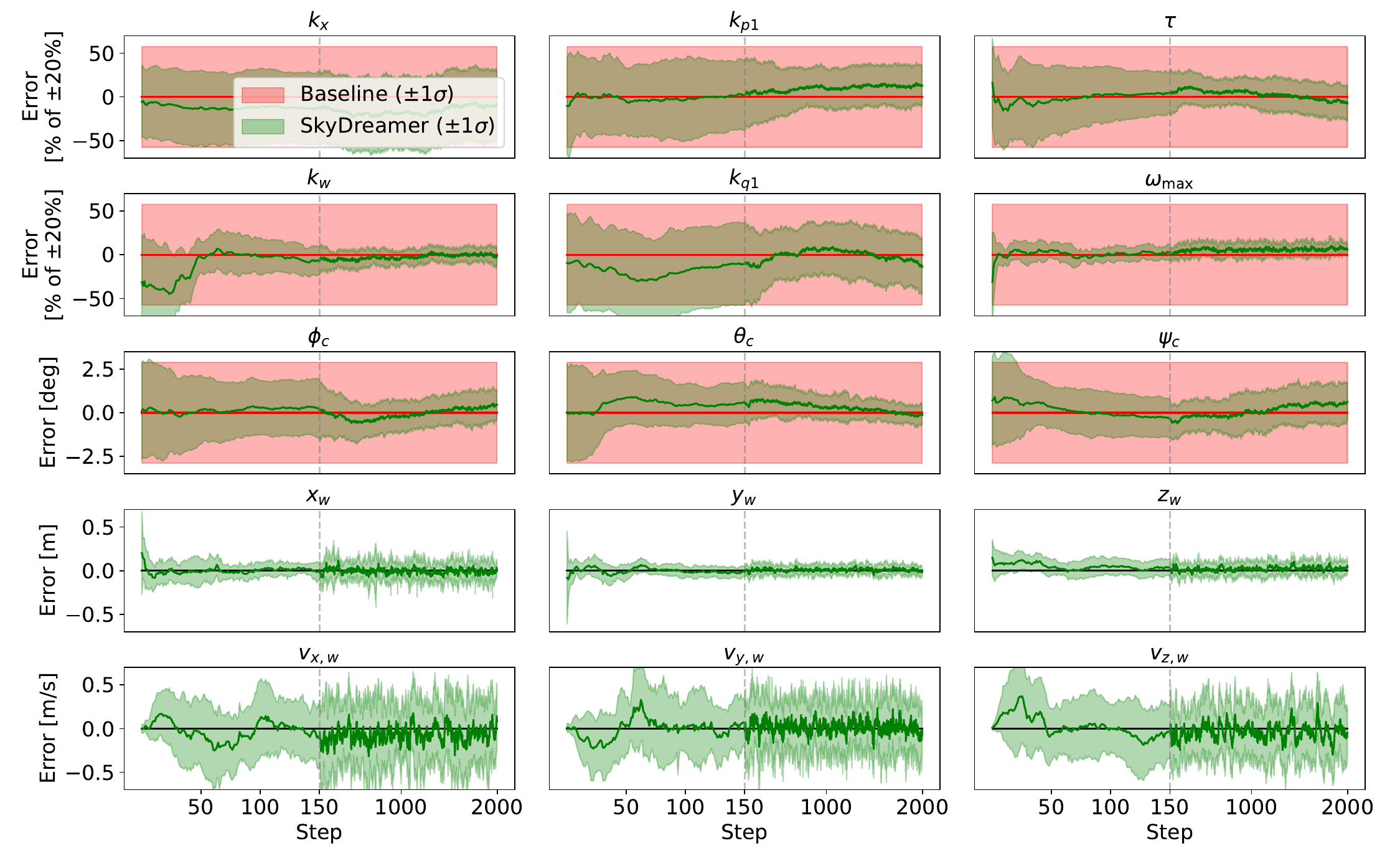}
    \caption{Selected SkyDreamer parameter, extrinsic, and state estimation over time for the ladder inverted loop track with orange gates. The left half of each plot shows the first 150 steps, highlighting the speed of convergence, while the right half shows the remaining 1850 steps, illustrating any drift over time. The red distribution represents the baseline, computed using the theoretical standard deviation of a uniform distribution representing the distribution of the training ranges in \autoref{tab:randomization_ranges}. The green distribution shows SkyDreamer’s estimation over time, obtained by aggregating results from 20 differently initialized drones that completed a successful flight over 2000 timesteps each, following the evaluation parameter ranges from \autoref{tab:randomization_ranges}. Distributions such as those for $k_x$ align with the baseline distribution, indicating that SkyDreamer failed to estimate these parameters. Other distributions are narrower, showing that SkyDreamer successfully estimated them on the fly. Parameters like $\omega_{\max}$ and $k_w$ converge quickly and remain stable over time, while parameters like $k_{p1}$ converge more slowly and show gradual drift over time. }
    \label{fig:all_in_one_sim_params}
\end{figure*}

We deploy the learned policy onboard the same drone as in Bahnam \textit{et al.} \cite{aibeatshumandeWagter2025} (also shown in \autoref{fig:overview_image}) on an \textit{NVIDIA Jetson Orin NX 16GB}, operating entirely autonomously without any external aid. The JAX model is first converted to PyTorch using our custom implementation, then exported to ONNX, and finally optimized and executed with TensorRT. The full policy, including encoder, sequence model, and actor, achieves an average runtime of $1.3\,\text{ms}$ in benchmark tests, while the segmentation model averages $3\,\text{ms}$. 

The onboard camera is a low-cost rolling shutter \textit{Arducam IMX219 Wide Angle Camera}\footnote{https://www.arducam.com/b0179-arducam-imx219-wide-angle-camera-module-for-nvidia-jetson-board.html} with a $175^\circ$ field of view operating at $90\,\text{Hz}$, which sets the control frequency of the policy. All timing is anchored to the camera timestamps. During training, we simulate an image delay of $33\,\text{ms}$ and an action delay of $11\,\text{ms}$ to match deployment conditions. Motor commands are sent immediately to the flight controller with a desired execution timestamp, and sufficient buffering is implemented to handle occasional timing variability in both the policy and segmentation model.

\begin{figure*}[!htb]
    \centering
    \begin{subfigure}[b]{0.67\linewidth}
        \centering
        \includegraphics[width=\linewidth]{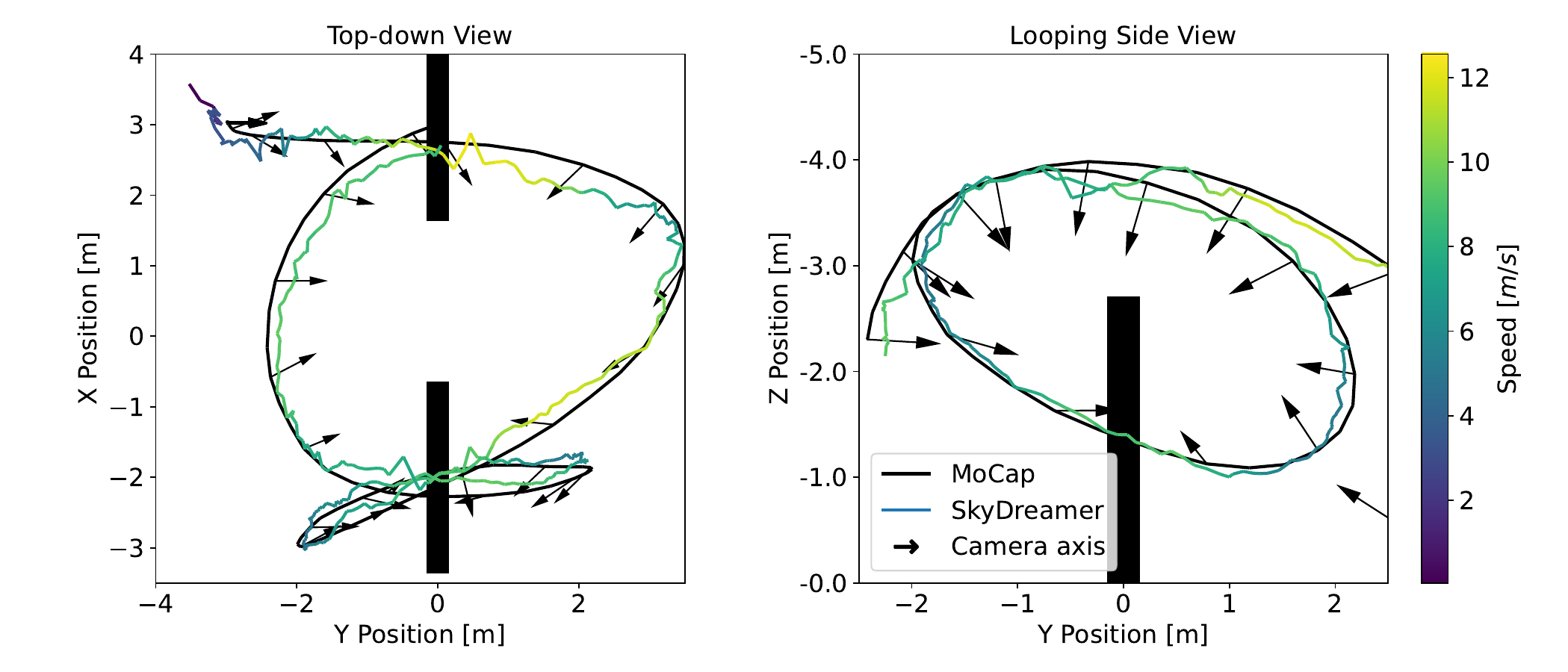}
    \end{subfigure}
    \hfill
    \begin{subfigure}[b]{0.32\linewidth}
        \centering
        \includegraphics[width=\linewidth,trim={5cm 0 0 0},clip]{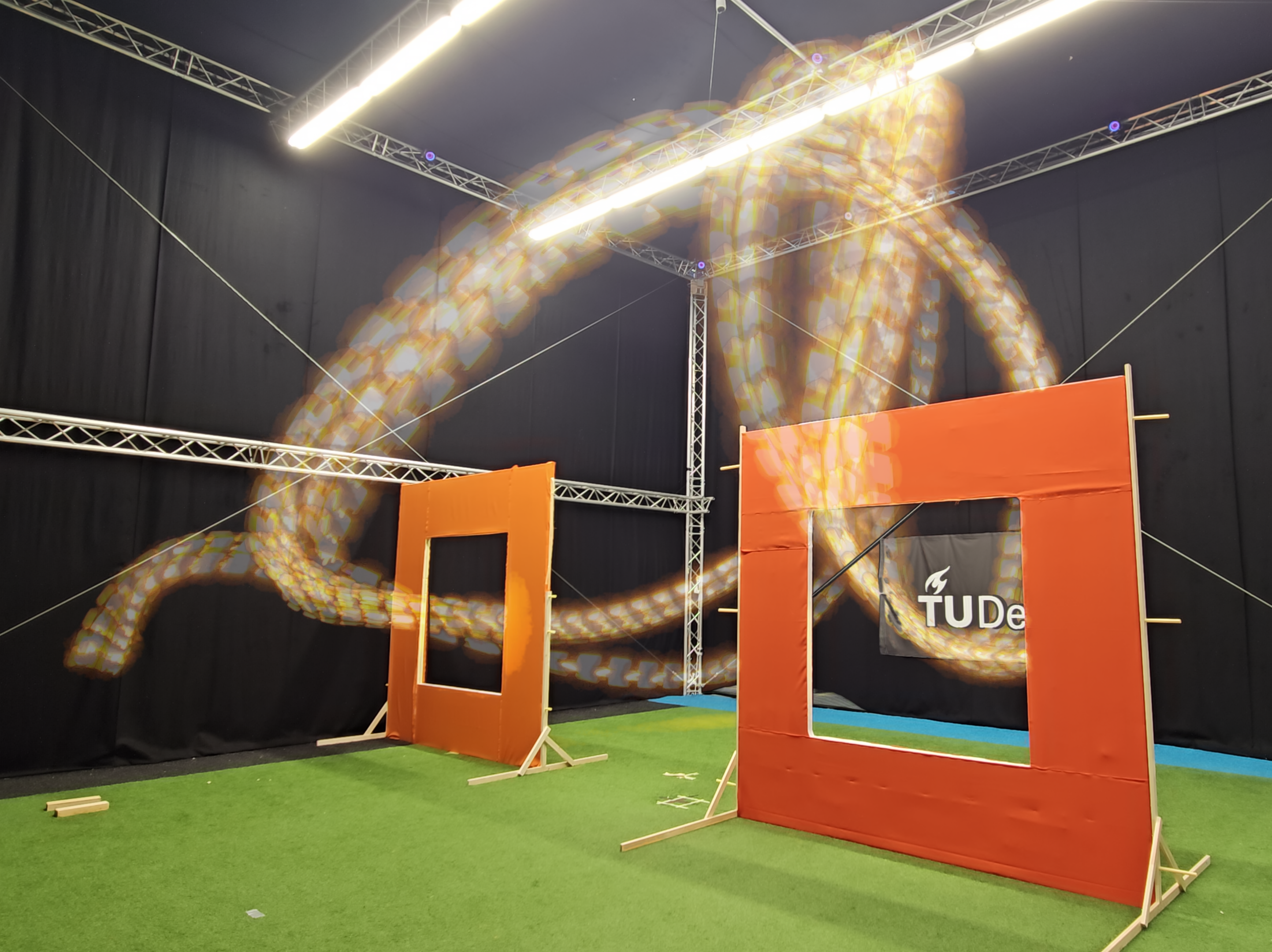}
    \end{subfigure}
    \caption{Real-world inverted loop track flown by SkyDreamer. The flight begins by passing through the first gate. SkyDreamer then flies over the second gate, performs a maneuver similar to a split-S through it, and flies back over the gate, completing the inverted loop. The lap is completed by passing through the first gate. In the plots, the black lines indicate ground-truth positions obtained using MoCap as a reference, which were not used by SkyDreamer. The colored lines show the position and velocity estimates of SkyDreamer, with color denoting speed. The black blocks mark the gate locations with exaggerated thickness. The black arrows indicate the camera principal axis direction, which show that SkyDreamer naturally orients its camera toward the gates: an emergent behavior achieved without an explicit perception reward. The right image shows a composite image of five laps flown by SkyDreamer in the real world. The overlaid trajectories of all five laps converge at the center points of the gates, visually demonstrating that SkyDreamer consistently flies through the center of each gate.}
    \label{fig:inverted_looping_real_combined_orange}
\end{figure*}

\begin{figure*}[!htb]
    \centering
    \begin{subfigure}[t]{0.67\linewidth}
        \centering
        \includegraphics[width=\linewidth]{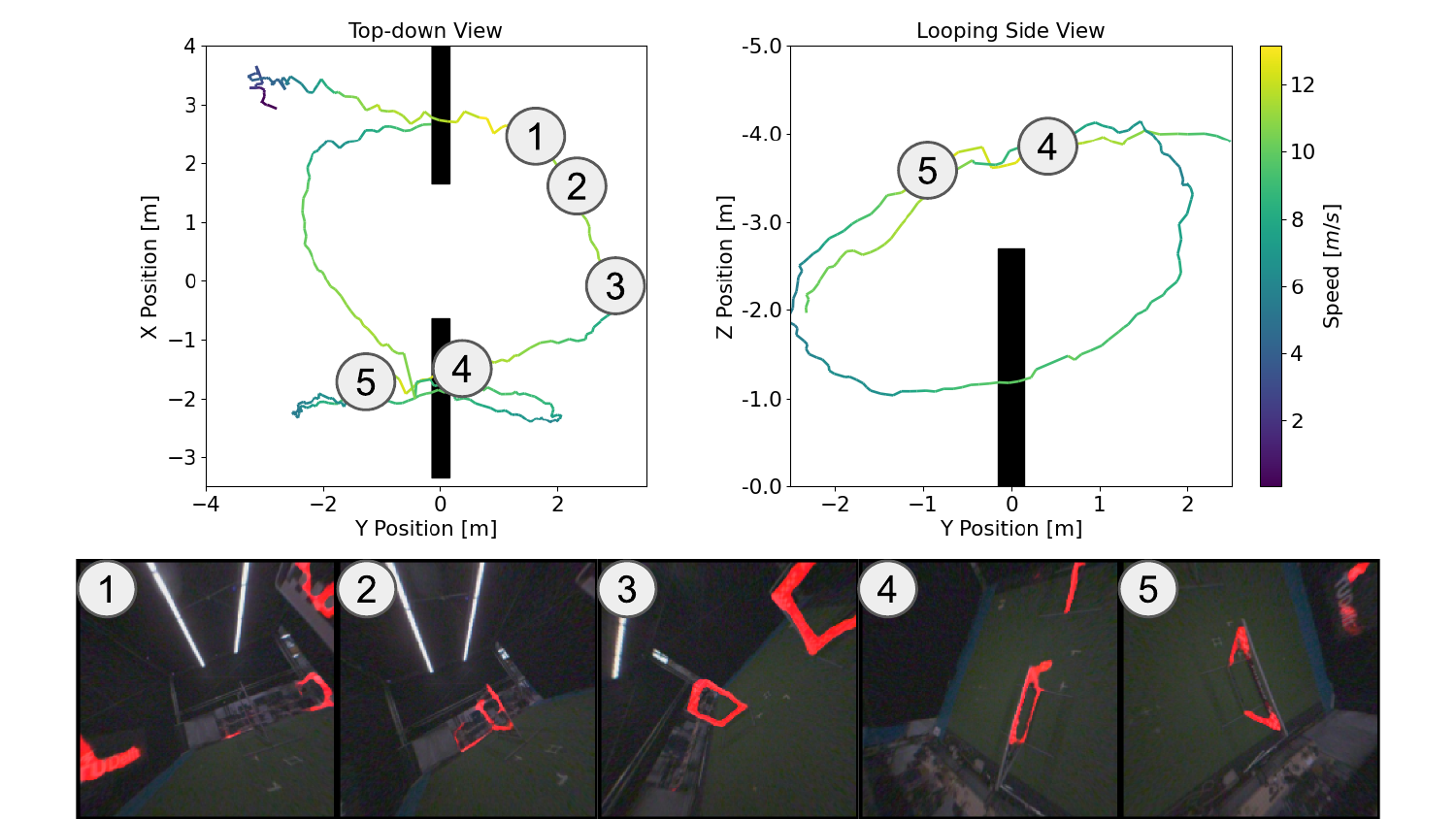}
    \end{subfigure}
    \hfill
    \begin{subfigure}[t]{0.32\linewidth}
        \centering
        \includegraphics[width=\linewidth,trim={5cm 0 0 0},clip]{figures/drone_trajectory_mavlab_invloop.png}
    \end{subfigure}
    \caption{Real-world inverted loop track with MAVLab gates flown by SkyDreamer. The flight begins by passing through the first gate. SkyDreamer then flies over the second gate, performs a maneuver similar to a split-S through it, and flies back over the gate, completing the inverted loop. The lap is completed by passing through the first gate. In the plots, the colored lines show the position and velocity estimates of SkyDreamer, with color denoting speed. The black blocks mark the gate locations with exaggerated thickness. Below the plots, selected suboptimal segmentations produced by GateNet are shown, with their corresponding locations indicated by the numbers. Red pixels denote regions classified as gates by GateNet. MoCap data could not be recorded during this flight due to technical issues, so ground-truth positions are unavailable. The right image shows a composite image of one lap flown by SkyDreamer in the real world.}
    \label{fig:inverted_looping_real_combined_mavlab}
\end{figure*}

Two types of gates are used in our experiments: plain square orange gates with an inner size of 1.5m and an outer size of 2.7m shown in \autoref{fig:inverted_looping_real_combined_orange}, similar to those in Bahnam \textit{et al.} \cite{aibeatshumandeWagter2025}, and MAVLab gates, which resemble those in De Wagter \textit{et al.} \cite{dewagteralphapilot2019}, with an inner size of 1.5m and an outer size of 2.1m shown in \autoref{fig:inverted_looping_real_combined_mavlab}.

For the orange gates, we train GateNet using only 200 manually labeled real-world images, which we find sufficient to achieve near-perfect segmentations. In contrast, the MAVLab gates are more challenging to segment due to their dark blue and black colors, thinner structure, the presence of logos, and the similarity of background textures at one side of the testing area (e.g., nearby blue machinery). For this case, we train on 700 manually labeled real-world images combined with 8,500 synthetically generated examples, following the synthetic generation procedure described in Bahnam \textit{et al.} \cite{aibeatshumandeWagter2025}. Importantly, our objective is not to optimize GateNet for peak segmentation accuracy, as illustrated by some deliberately retained suboptimal examples in \autoref{fig:inverted_looping_real_combined_mavlab}, but rather to demonstrate that SkyDreamer can successfully cope with poor-quality visual inputs and demonstrate a non-trivial visual sim-to-real transfer.

\subsection{Simulation}

In the simulation experiments, we evaluate state and parameter estimation, demonstrate interpretability and an important aspect of vision-based flight: perceptual behavior. To this end, we investigate SkyDreamer's performance on the ladder inverted loop track with orange gates, shown in \autoref{fig:perception_ladder_sim_combined}. This track is trained with a smaller tunnel size $t_g = 0.3m$, to further demonstrate SkyDreamer's ability to execute tight maneuvers.

\autoref{fig:perception_ladder_sim_combined} provides a qualitative analysis of SkyDreamer's flight and state estimation behavior. SkyDreamer's position estimate aligns closely with the ground-truth position, indicating accurate state estimation. However, the estimated states exhibit small, high-frequency jumps around the ground truth. We hypothesize that this behavior may arise from the discretized nature of the input representations $\mathbf{z}_t$. Another possible contributing factor is that the world model may not fully capture the underlying physics, as observed in similar settings \cite{vafa2025foundationmodelfoundusing}.

The maneuvers themselves are tight: SkyDreamer completes the ladder while remaining mostly within $1 \, \text{m}$ of the gate and executes the inverted loop in an almost perfect circle with a radius of roughly $1.5 \, \text{m}$, closely matching the separation between the actual gate and the virtual gate of $2.7\, \text{m}$. Speeds reach up to $13\, \text{m/s}$ and accelerations up to $6\, \text{g}$ within a confined area of just $6 \times 4\, \text{m}$, demonstrating both agility and high-speed flight.

The tightness and speed of these maneuvers do not come at the expense of perceptual behavior. The arrows in \autoref{fig:perception_ladder_sim_combined} indicating the camera principal axis show that SkyDreamer consistently focuses on the gates to maximize gate visibility, despite the absence of a perception reward. This emergent behavior, also observed in Romero \textit{et al.} \cite{romero2025dreamflymodelbasedreinforcement}, is thus evident even in more complex maneuvers such as the ladder and inverted loop.

\autoref{fig:all_in_one_sim_params} shows SkyDreamer's state and parameter estimation over time. Some dynamical parameters are estimated better than others: while drag ($k_x$) is difficult to estimate, performance is moderate for individual propeller responses in roll ($k_{p1}$) and pitch ($k_{q1}$). Parameters such as yaw effectiveness (e.g., $k_{r1}$) are not accurately estimated.

However, SkyDreamer excels at identifying two parameters that are crucial for thrust response: the thrust effectiveness $k_w$ and maximum motor RPM $\omega_{\max}$. Correct estimation of these parameters enables SkyDreamer to perform high-speed maneuvers and tight cornering tailored to the specific drone despite domain randomization, similar to Origer \textit{et al.} \cite{origer2023guidancecontrolnetworks}. Other parameters, such as the motor response $\tau$, are also estimated accurately. 

Overall, parameters that have a direct and strong effect on the drone’s inputs, like $\omega_{\text{max}}$, are estimated most accurately, whereas parameters with smaller long-term effects or those affecting only a single motor are more challenging to estimate.

In addition to its final flight performance, the estimated parameters converge quickly within the first $50$–$100$ steps ($0.6\,\text{s}$ to $1.1\,\text{s}$) -- before the drone passes through the first gate -- enabling SkyDreamer to fly safely and at high speed from the start. Some parameters, however, exhibit drift in the mean of their distributions over time, indicating that the sequence length used during world model learning ($256$ steps) does not necessarily generalize to the longer sequences encountered during flight ($2000$ steps). It should however be noted that the accuracy of parameter identification depends on the training run: some runs yield more precise estimates, while others exhibit lower accuracy and can show significant drift over time, without resulting in a significant loss of flight performance.

SkyDreamer also estimates the camera extrinsics over time. Convergence is slightly slower than for some dynamical parameters, but it reaches a standard deviation of roughly $1^\circ$ over time, indicating accurate on the fly estimation of the camera's extrinsics. Notably, the pitch angle of the camera is estimated most accurately, while the roll angle is slightly less accurate, though the difference is marginal.

Finally, we evaluate the estimation of position and velocity over time -- two of the most important yet challenging state variables for classical methods \cite{aibeatshumandeWagter2025}. Despite randomized initial positions, the estimates converge quickly, within at most 10 steps, to the nominal distribution, with a typical standard deviation of only $10-15\,\text{cm}$ across the randomized drones, remaining centered over time. Velocity estimates are also accurate, with a standard deviation of approximately $0.5\,\text{m/s}$ in all axes, and remain centered around zero, indicating no drift over time.

\subsection{Real-world small tracks}

\begin{figure}[!ht]
    \centering
    \includegraphics[width=0.48\textwidth]{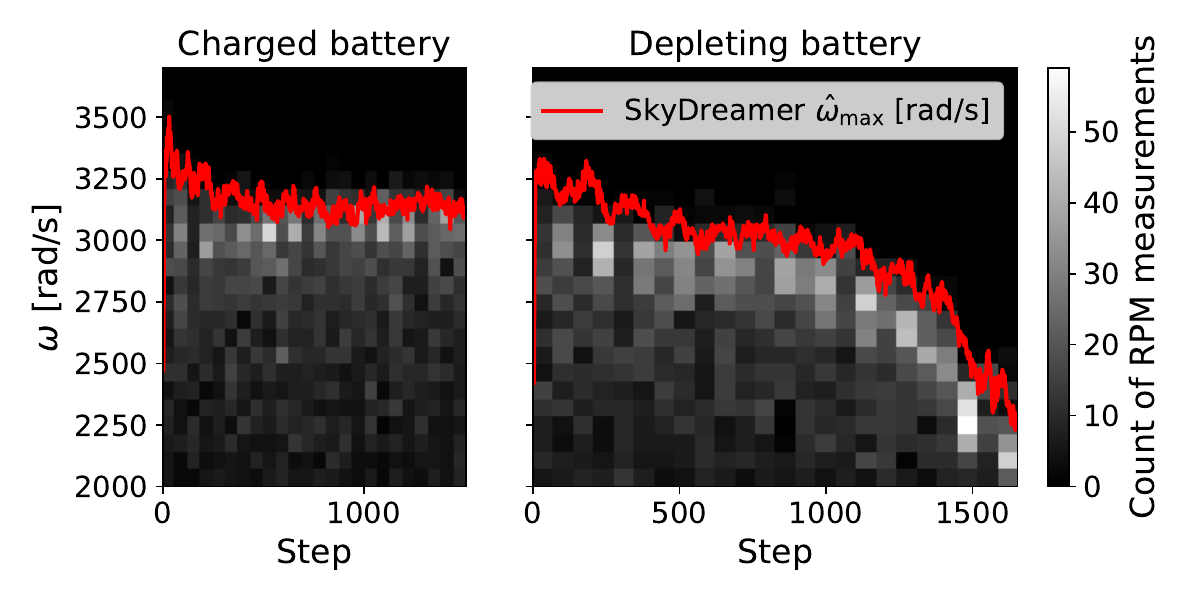}
    \caption{Real-world battery degradation over five laps of the inverted loop track with orange gates. The heatmap shows measured motor RPMs over time, where brighter regions indicate higher occurrence at a given timestamp. The red line denotes the maximum RPM estimated by SkyDreamer, $\hat{\omega}_{\max}$. The left-hand plot indicates a fully charged battery which is not excessively discharged, while the right-hand plot depicts a battery undergoing excessive discharge, resulting in reduced RPM toward the end of the flight. However, SkyDreamer accounts for this, correctly estimates $\hat{\omega}_{\max}$, adjusts its flight path accordingly and successfully completes all five laps with inverted loops.}
    \label{fig:battery_comparison}
\end{figure}

In the first real-world experiments, we evaluate SkyDreamer's sim-to-real transfer, robustness, and performance across three small tracks: the inverted loop with both orange and MAVLab gates (\autoref{fig:inverted_looping_real_combined_orange} and \autoref{fig:inverted_looping_real_combined_mavlab}), and the ladder inverted loop track with orange gates, which uses the model from the same training run as the simulation experiments (\autoref{fig:perception_ladder_sim_combined}).

\autoref{tab:lap_times} summarizes SkyDreamer’s performance on these tracks. For each track, we conduct five flights of five laps each, resulting in $25$ laps per track and $75$ laps in total. SkyDreamer successfully completes all laps without crashing. For the orange inverted loop track, one flight was performed with a degrading battery, leading to a higher average lap time and a larger standard deviation. Nevertheless, SkyDreamer still successfully completed all five laps. During these flights, SkyDreamer reached measured accelerations of up to $6\, \text{g}$ and estimated speeds of up to $13\,\text{m/s}$, demonstrating impressive agility with flight areas typically confined to $6 \times 6\,\text{m}$.

\begin{table}[ht!]
\centering
\caption{Success rates and lap times for the real-world small tracks flown without external aid. All tracks achieved 100\% success over five flights, with five laps per flight. The $\pm$ indicates the standard deviation, which is higher for the inverted loop (orange) track due to one flight with a partially charged battery. One flight on the ladder loop track was excluded because the flight plan condition $\hat{\mathbf{x}}_g > 0.0\,\text{m}$ never triggered, resulting in an aborted flight; the condition was subsequently adjusted to $\hat{\mathbf{x}}_g > -0.15\,\text{m}$. For the MAVLab gates, a part of the environment resembling a gate had to be covered, as GateNet consistently misclassified it.}
\label{tab:lap_times}
\begin{tabular}{@{}lccc@{}}
\hline
\textbf{Metric} & 
\shortstack{\textbf{Loop}\\ \\(orange)} & 
\shortstack{\textbf{Ladder}\\\textbf{loop}\\(orange)} & 
\shortstack{\textbf{Loop}\\ \\(MAVLab)} \\
\hline
Success rate & 100\% & 100\% & 100\% \\
Number of flights & 5 & 5 & 5 \\
Laps per flight & 5 & 5 & 5 \\
From start [s] & 17.03 ± 0.76 & 18.87 ± 0.23 & 15.56 ± 0.20 \\
From first gate [s] & 16.36 ± 0.77 & 18.24 ± 0.22 & 14.88 ± 0.20 \\
Lap 1 [s] & 3.37 ± 0.08 & 3.78 ± 0.09 & 2.99 ± 0.07 \\
Lap 2–5 [s] & 3.25 ± 0.22 & 3.62 ± 0.06 & 2.97 ± 0.08 \\
\hline
\end{tabular}
\end{table}

To demonstrate SkyDreamer's parameter estimation in the real world, \autoref{fig:battery_comparison} illustrates its performance during flights with and without a fully charged battery. The figures show that the maximum RPM estimated by SkyDreamer, $\hat{\omega}_{\text{max}}$, closely follows the actual measured RPMs, even though it was never trained with non-constant parameters. By the final lap, the maximum RPM drops from 3200 rad/s to 2200 rad/s, a 30\% reduction, well outside the training randomization bounds of [2480, 3720] rad/s. Despite this, SkyDreamer detects the change, adapts its flight path, and still completes several inverted loops. This demonstrates that online parameter estimation allows SkyDreamer to adapt on the fly, maintaining both safety and agility despite a significant loss of maximum available thrust.

In addition to estimating parameters, \autoref{fig:inverted_looping_real_combined_orange} shows that the state estimation transfers well to the real world, where SkyDreamer's state estimates only show a slight offset to the MoCap positions in the $xy$-plane, and an almost perfect overlap in the $yz$-plane during the loop. The figure also reveals that SkyDreamer takes less tight loops compared to the ladder inverted loop track, which can be attributed to the larger tunnel size ($t_g = 0.8$ m), forcing it to take wider turns.

To visually illustrate SkyDreamer's flight path and consistency, \autoref{fig:inverted_looping_real_combined_orange} shows a composite image of five laps on the inverted loop track. Remarkably, the trajectories from all five laps converge closely to each other and to the center of the gates, demonstrating that SkyDreamer navigates both consistently and safely.

While SkyDreamer generalizes well with orange gates -- whose high contrast and uniform appearance yield near-perfect segmentation masks closely matching the non-augmented rendered ones -- we also evaluate it with MAVLab gates. For these, GateNet’s performance degrades due to poor lighting, low contrast, and distracting background elements such as machinery that visually resemble gates. As shown in \autoref{fig:inverted_looping_real_combined_mavlab}, GateNet occasionally misclassifies background regions as gates, produces rounded masks, and often fails to fully align with the actual gate boundaries. However, rather than adopting more advanced segmentation models like the Swin Transformer V2 \cite{swin_transformer_v2} to improve the segmentation performance -- as done in Geles \textit{et al.} \cite{geles2024demonstratingagileflightpixels} -- we intentionally retain GateNet’s baseline performance to demonstrate that SkyDreamer can operate in the real world with poor-quality perception.

\begin{figure*}[!ht]
    \centering
    \begin{subfigure}[b]{0.52\linewidth}
        \centering
        \includegraphics[width=\linewidth]{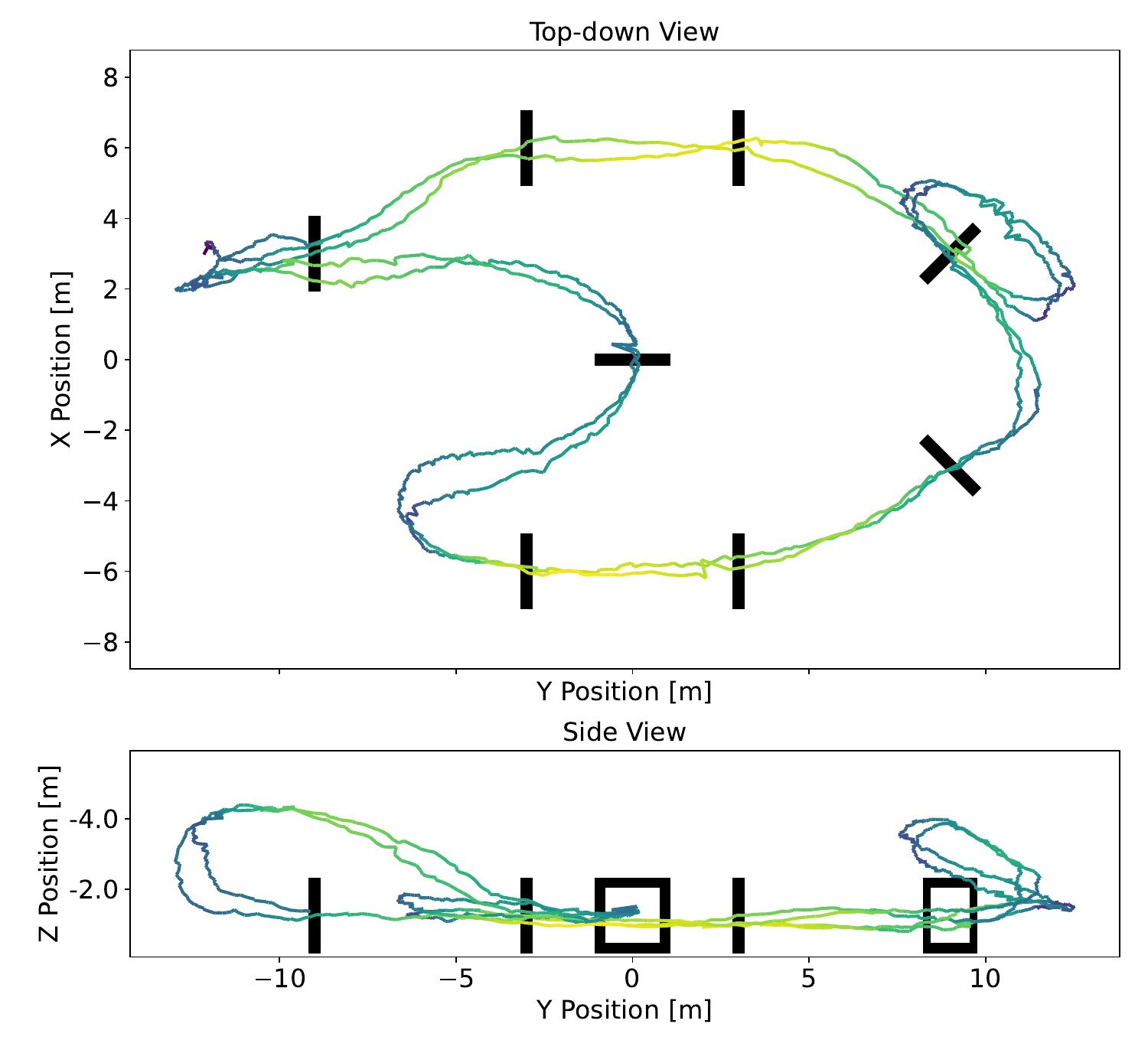}
    \end{subfigure}
    \hfill
    \begin{subfigure}[b]{0.47\linewidth}
        \centering
        \begin{subfigure}[t]{0.98\linewidth}
            \centering
            \includegraphics[width=0.98\linewidth]{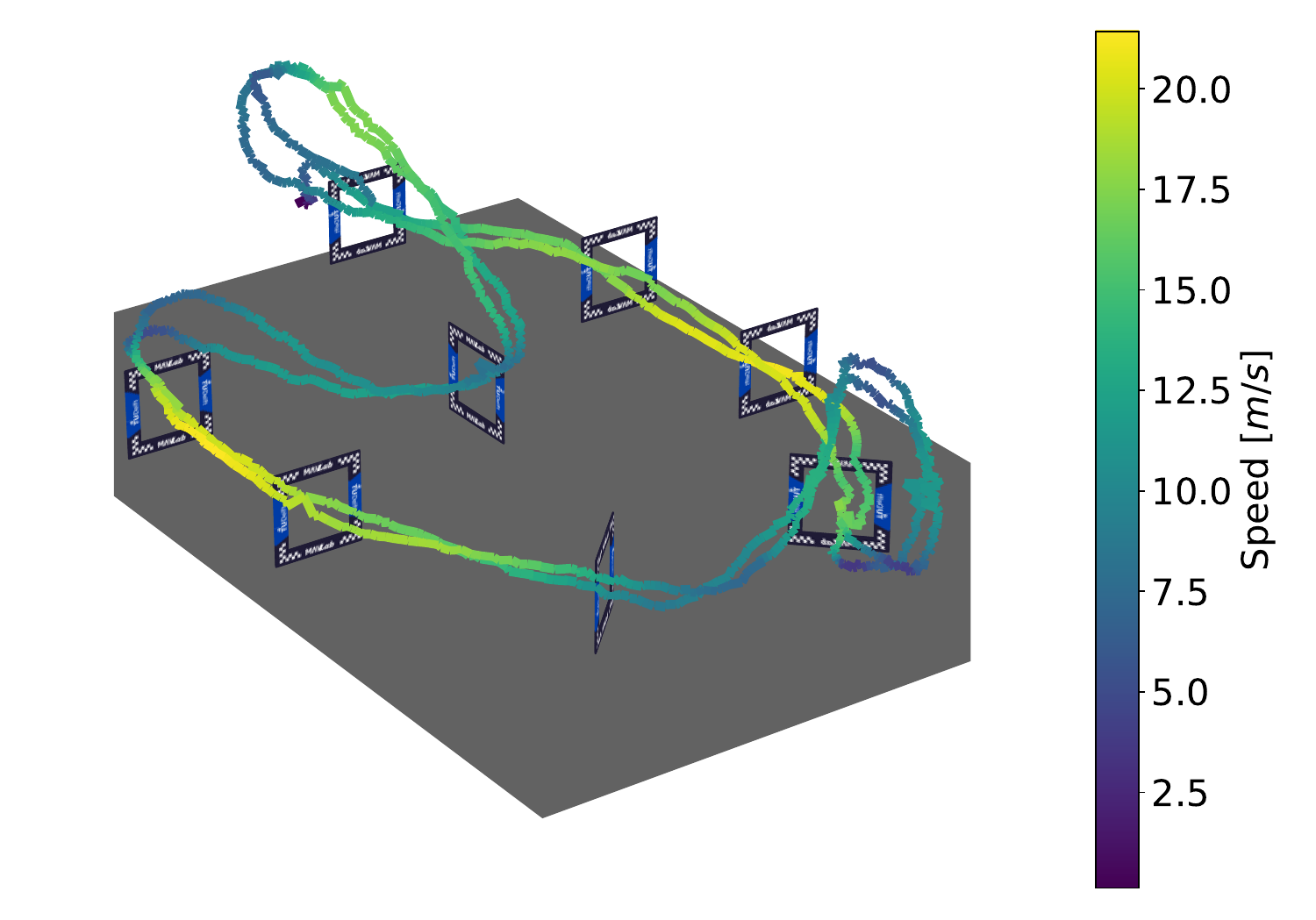}
        \end{subfigure}
        \vfill
        \begin{subfigure}[t]{0.98\linewidth}
            \centering
            \includegraphics[width=0.98\linewidth]{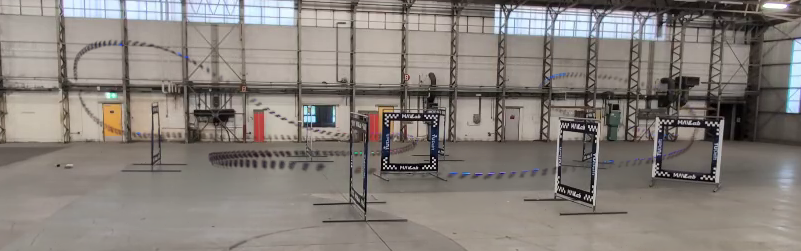}
        \end{subfigure}
    \end{subfigure}
    \caption{Real-world big track flown by SkyDreamer. In the top-down view, the flight begins in front of the top-left gate. After the first gate, the trajectory continues with a slight left turn, two subsequent gates, a right turn, and a ladder maneuver in which the drone passes through the lower gate, performs a full 360° left turn, and flies back over it. From there, SkyDreamer extends the right turn, executing a steep dive into the next gate, and continues through several more gates before performing a tight braking maneuver to initiate a sharp right turn. After the following gate, the track concludes with a split-S maneuver over the first gate. In the plots, the colored lines show SkyDreamer’s position and velocity estimates for two real-world laps, with color indicating speed. Black blocks mark gate locations with exaggerated thickness. The bottom-right image presents a composite visualization of one lap from another flight flown by SkyDreamer.}
    \label{fig:perception_uvv_combined}
\end{figure*}

Thanks to the use of StochGAN and additional data augmentations, SkyDreamer still completes 25 consecutive laps on the inverted loop track with MAVLab gates, as summarized in \autoref{tab:lap_times}. Despite GateNet's poor performance, SkyDreamer demonstrates robustness to perceptual errors and bridges a non-trivial visual reality gap. This opens the door to using alternative more generalizable pixel-level abstractions such as depth maps, which could potentially enable high-speed, agile flight in unstructured environments.

\subsection{Real-world big track}

Additional real-world experiments demonstrate SkyDreamer’s ability to fly the big track with MAVLab gates illustrated in \autoref{fig:perception_uvv_combined}. This track combines high-speed flight with challenging maneuvers, including a split-S, a sharp right turn, and a ladder. The experiments are conducted in the real world in a larger hall and under poorer lighting conditions, further highlighting SkyDreamer’s robustness across environments.

We train SkyDreamer with slightly modified settings for these experiments: we disable symlog scaling in both the encoder and decoder \cite{dreamerv3_nature}, set the \texttt{train\_ratio} to $64$, and train for a total of $35$ million environment steps. The second training stage begins after $23$ million steps, and the third stage after $31$ million steps. Furthermore, we set $t_g = 0.5$ and introduce disturbances of $\pm 300\,\text{rad/s}$ to $\omega_{max}$, randomly resampled every 10 timesteps, to account for imperfect actuator response modeling.

During these experiments, five out of six flights consisting of two laps each were successful. The only crash occurred with a drone whose extrinsics lay outside the training distribution specified in \autoref{tab:randomization_ranges}, as verified by a preflight Kalibr calibration, likely explaining the failure. The five successful flights were all with the same drone whose extrinsics are within the training distribution. In three of these flights, minor gate contact was audible once or twice, but SkyDreamer consistently carried on successfully, highlighting its ability to withstand and adapt to contact forces it was not explicitly trained for. In this track, SkyDreamer reaches accelerations comparable to those on the smaller tracks (around $6 \, \text{g}$) while achieving significantly higher estimated speeds of up to $21 \, \text{m/s}$. These results highlight SkyDreamer's ability to combine high-speed flight with agile and tight maneuvers.

\section{Conclusion}

In this work, we presented SkyDreamer, to the best of our knowledge the first end-to-end vision-based ADR policy mapping pixel-level representations directly to motor commands. By building on informed Dreamer and extending it to end-to-end vision-based ADR, SkyDreamer provides interpretability by decoding to states and drone-specific parameters, effectively turning the world model into an implicit state and parameter estimator. SkyDreamer runs fully onboard without external aid, resolves visual ambiguities by tracking progress through the world model’s decoded state, and requires no extrinsic camera calibration, enabling rapid deployment across different drones without retraining. 

Our real-world experiments demonstrate non-trivial sim-to-real transfer despite operating on low-quality segmentation masks, reliable performance across diverse tracks, and robustness to battery depletion. SkyDreamer successfully executes tight maneuvers and reaches speeds up to $21 \, \text{m/s}$ with accelerations of up to $6 \, \text{g}$ -- speed and agility not previously demonstrated by other end-to-end vision-based ADR methods in the real world. These results highlight SkyDreamer’s adaptability to important aspects of the reality gap, bringing robustness while still achieving extremely high-speed, agile flight.

Despite these promising results, several limitations remain. Parameter estimates tend to drift over time, and their quality varies across training runs. Decoded state estimates jump between timesteps, and SkyDreamer is still vulnerable to false positives in segmentation masks. In addition, training remains computationally demanding, requiring roughly $50$ hours to complete.

Future work will explore extending SkyDreamer to more generalizable visual inputs, such as depth maps, flying unseen tracks, and potentially generalizing to unstructured environments or hybrid tasks by combining drone racing with obstacle avoidance. Beyond racing, SkyDreamer represents a promising step toward deploying end-to-end vision-based yet interpretable policies for high-speed, agile flight in real-world settings.

\section*{Acknowledgments}
We are grateful to the other members of the drone racing team -- Anton Lang, Till Blaha, Quentin Missine and Erin Lucassen -- for their valuable insights and support during system integration and flight experiments.

\bibliographystyle{IEEEtran}
\bibliography{skydreamer_bibliography}

\begin{appendices}

\section{GateNet Implementation Details}\label{app:gatenet}
\paragraph{Network}
For segmentation, we adopt a U-Net style architecture \cite{unet_original}, referred to as GateNet, similar to Bahnam \textit{et al.} \cite{aibeatshumandeWagter2025}. The network consists of an encoder–decoder with skip connections from the encoder to the corresponding decoder layer, following the standard U-Net design \cite{unet_original}, where channel dimensions are scaled by a factor $f$. Each block consists of double convolutional layers with batch normalization and ReLU activation functions.  

Let \texttt{inc} denote the initial double convolutional block. \texttt{downk} denotes a max-pooling layer followed by a double convolutional block with $k$ output channels. Before applying \texttt{downk}, the corresponding features are stored as skip connections for the decoder. \texttt{upk} denotes a transposed convolution and batch normalization, followed by adding the skip connections from the corresponding encoder layer, and then a double convolutional block with $k$ output channels. Finally, \texttt{outc-k} denotes a $1 \times 1$ convolution mapping to a single output channel and a sigmoid activation function.

The resulting GateNet architecture is:  

\begin{align*}
\begin{array}{c c c}
\text{Encoder} & \text{Decoder} & \text{Outputs} \\[2mm]
\texttt{inc-64/f} & \rightarrow \texttt{up4-64/f} & \rightarrow \texttt{outc4-1} \\[1mm]
\downarrow & \uparrow & \\[1mm]
\texttt{down1-128/f} & \rightarrow \texttt{up3-64/f} & \rightarrow \texttt{outc3-1} \\[1mm]
\downarrow & \uparrow & \\[1mm]
\texttt{down2-256/f} & \rightarrow \texttt{up2-128/f} & \rightarrow \texttt{outc2-1} \\[1mm]
\downarrow & \uparrow & \\[1mm]
\texttt{down3-512/f} & \rightarrow \texttt{up1-256/f} & \rightarrow \texttt{outc1-1} \\[1mm]
\downarrow & \nearrow & \\[1mm]
\texttt{down4-512/f} & & \rightarrow \texttt{outc0-1} \\
\end{array}
\end{align*}

The network produces five output maps $\{y_0, y_1, y_2, y_3, y_4\}$ at different resolutions. These multi-scale predictions are supervised with auxiliary losses to improve gradient flow and overall segmentation accuracy.  

\paragraph{Training setup}
We use the same hyperparameters as in Bahnam \textit{et al.} \cite{aibeatshumandeWagter2025}. Each output map is supervised with a combination of Dice loss and binary cross-entropy (BCE) loss:  

\begin{align*}
\mathcal{L}_i = \mathcal{L}_{\text{Dice}}(y_i, \hat{y}_i) + 2 \cdot \mathcal{L}_{\text{BCE}}(y_i, \hat{y}_i),
\end{align*}

and we apply output-specific scaling factors to emphasize higher-resolution predictions:

\begin{align*}
\mathcal{L}_{\text{total}} = 4 \cdot \mathcal{L}_0 + 2 \cdot \mathcal{L}_1 + \sum_{i=2}^{4} \mathcal{L}_i.
\end{align*}

All convolutional layers are initialized using Xavier uniform initialization \cite{xavier_init}.

\paragraph{Data augmentation}
We largely use the same augmentations as in Bahnam \textit{et al.} \cite{aibeatshumandeWagter2025}. Since our images are captured with a low exposure time (1~ms), we do not simulate motion blur with KernelBlur. Instead, we apply stronger shot noise (standard deviation of 40 for pixel values in the range 0--255 instead of 25) to mimic the increased noise associated with low exposure times.

\paragraph{Gate-specific implementation}
For MAVLab gates, we use a resolution of $196 \times 196$ with $f = 2$. For orange gates, we use a resolution of $384 \times 384$ with $f = 4$.

\section{StochGAN Implementation Details}\label{app:stochgan}
\paragraph{Generator}
The generator is a CNN-based encoder–residual–decoder network, largely based on Zhu \textit{et al.} \cite{cyclegan}. We use 6 residual blocks and images of size $64 \times 64$, the same resolution as those rendered and used by SkyDreamer during training. Following a similar convention to Zhu \textit{et al.} \cite{cyclegan}, let \texttt{c7s1-k} denote a $7 \times 7$ Convolution–InstanceNorm–ReLU layer with $k$ filters and stride 1, using reflection padding. \texttt{dk} corresponds to a $3 \times 3$ Convolution–InstanceNorm–ReLU layer with $k$ filters and stride 2 for downsampling. \texttt{Rk} denotes a residual block consisting of two $3 \times 3$ Convolution–InstanceNorm–ReLU layers with $k$ filters, using reflection padding and a skip connection. \texttt{uk} represents a $3 \times 3$ transposed Convolution–InstanceNorm–ReLU layer with $k$ filters, stride 2 for upsampling, and reflection padding. Finally, the output layer is denoted as \texttt{c7s1-k-tanh}, which follows the same convention as \texttt{c7s1-k} but uses a \texttt{tanh} activation at the final layer.  

The resulting generator architecture is:  

\begin{align*}
\text{Encoder: } & \texttt{c7s1-32}, \; \texttt{d64}, \; \texttt{d128}\\ 
\text{Residual blocks: } & 6 \times \texttt{R128} \\
\text{Decoder: } & \texttt{u64}, \; \texttt{u32}, \; \texttt{c7s1-1-tanh}
\end{align*}

To obtain a StochGAN \cite{stochastic_gan}, we add an additional noise channel, uniformly sampled from $[-1,1]$, to the input image.

\paragraph{Discriminator}
Similar to Zhu \textit{et al.} \cite{cyclegan}, the discriminator follows a PatchGAN design \cite{patchgan}. Largely following the design and conventions in Zhu \textit{et al.} \cite{cyclegan}, let \texttt{Ck} denote a $4 \times 4$ Convolution–InstanceNorm–LeakyReLU layer with $k$ filters and stride 2. InstanceNorm is omitted in the first \texttt{Ck} layer, and LeakyReLU activations with a negative slope of 0.2 are used. After the last block, we apply zero-padding followed by a $4 \times 4$ convolution with 1 channel (\texttt{conv4-1}), producing a patch-wise realism score.  

The resulting discriminator architecture is:  
\[
\texttt{C32}, \; \texttt{C64}, \; \texttt{C128}, \; \texttt{C256}, \; \texttt{conv4-1}
\]

\paragraph{Training setup.}
We largely follow the training procedure of Zhu \textit{et al.} \cite{cyclegan}, except we use the Adam optimizer with $\beta_1 = 0.5$, $\beta_2 = 0.999$, and a base learning rate of $1.5 \times 10^{-4}$.

\end{appendices}

\end{document}